\documentclass[runningheads]{llncs}
\usepackage{graphicx}

\usepackage[dvipsnames]{xcolor}
\usepackage{tikz}
\usepackage{comment}
\usepackage{amsmath,amssymb} 
\usepackage{color}

\usepackage[accsupp]{axessibility}  

\usepackage[caption=false]{subfig}
\usepackage{graphicx}
\usepackage{amsmath}
\usepackage{amssymb}
\usepackage{booktabs}
\usepackage{arydshln}

\newcommand{\ie}{\emph{i.e.}~}
\newcommand{\eg}{\emph{e.g.}~}

\begin{document}
\pagestyle{headings}
\mainmatter
\def\ECCVSubNumber{4584}  

\title{Vector Quantized Image-to-Image Translation} 

\titlerunning{Vector Quantized Image-to-Image Translation}
%
\author{Yu-Jie Chen\protect\footnotemark[1]\inst{1,2}, Shin-I Cheng\protect\footnotemark[1]\inst{1,2} 
\and
Wei-Chen Chiu\inst{1,2} \and\\
Hung-Yu Tseng\inst{3} \and
Hsin-Ying Lee\inst{4}
}
\authorrunning{Y. Chen, S. Cheng et al.}
%
\institute{
$^1$National Chiao Tung University, Taiwan 
$^2$MediaTek-NCTU Research Center
$^3$Meta
$^4$Snap Inc.
}
\maketitle

\renewcommand{\thefootnote}{\fnsymbol{footnote}}
\footnotetext[1]{Equal contribution.
\newline Project page: \href{https://cyj407.github.io/VQ-I2I/}{https://cyj407.github.io/VQ-I2I/}}

\begin{figure}
    \includegraphics[width=\linewidth]{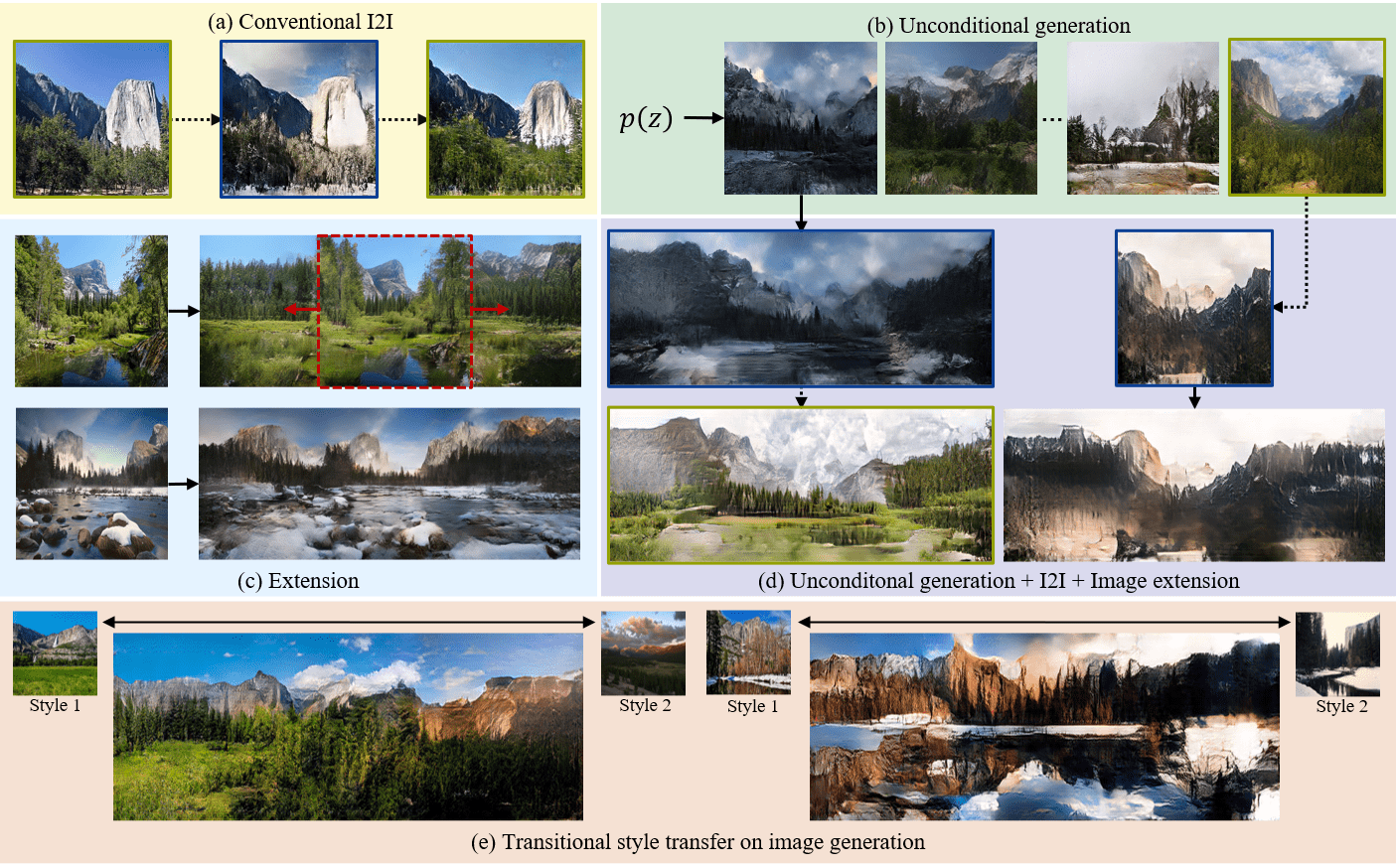}
    \centering
    \caption{
    \textbf{Applications of Vector Quantized Image-to-Image Translation.} Our proposed method enables several applications: (a) conventional image-to-image translation, (b) unconditional image generation, (c) image extension, (d) arbitrary combination of aforementioned operations, \eg translation and extension on unconditionally generated images, and (e) image generation with transitional stylization. Here we use \textcolor{OliveGreen}{green frame for summer images} and \textcolor{RoyalBlue}{blue frame for winter images}.
    }
    \label{fig:teaser}
\end{figure}
\begin{abstract}
Current image-to-image translation methods formulate the task with conditional generation models, leading to learning only the recolorization or regional changes as being constrained by the rich structural information provided by the conditional contexts.
In this work, we propose introducing the vector quantization technique into the image-to-image translation framework.
The vector quantized content representation can facilitate not only the translation, but also the unconditional  distribution shared among different domains.
Meanwhile, along with the disentangled style representation, the proposed method further enables the capability of image extension with flexibility in both intra- and inter-domains.
Qualitative and quantitative experiments demonstrate that our framework achieves comparable performance to the state-of-the-art image-to-image translation and image extension methods.
Compared to methods for individual tasks, the proposed method, as a unified framework, unleashes applications combining image-to-image translation, unconditional generation, and image extension altogether.
For example, it provides style variability for image generation and extension, and equips image-to-image translation with further extension capabilities.
\keywords{Image-to-Image Translation, Vector Quantization, Image Synthesis, Generative Models}
\end{abstract}

\section{Introduction}
\label{sec:intro}
Image-to-image translation (I2I) aims to learn the mapping between different visual domains.
Upon being formulated as a conditional generation problem, I2I methods can tackle translation with either paired~\cite{isola2017image} or unpaired data~\cite{zhu2017unpaired}, and perform diverse translations by disentangling the content and style factors of each input domain~\cite{huang2018multimodal,lee2018diverse,zhu2017toward}.
These I2I methods unleash various applications, such as style transfer~\cite{huang2017arbitrary}, synthesis from semantic map or layout~\cite{cheng2020controllable,huang2020semantic,park2019semantic,tseng2020retrievegan}, domain adaptation~\cite{deng2018image,murez2018image}, and super-resolution~\cite{ledig2017photo}.

Most existing I2I methods model the task as a pixel-level conditional generation problem.
However, as the conditional contexts are already informative in structure and details, the translation tends to learn simple recolorization  or regional transformation without understanding the real target distribution. 
%
Is it possible to jointly learn the translation as well as the unconditional distribution to fully exploit the data and make both trainings mutually beneficiary?
One intuitive formulation is to define a domain-invariant joint latent distribution, then perform domain-specific maximum likelihood estimation on it.
Pixel space is a natural option for the joint latent distribution, yet it struggles to scale due to its computational expensive auto-regressive process.
Recently, vector quantization (VQ) technique has shown its effectiveness as an intermediate representation of generative models~\cite{esser2021taming,razavi2019generating}.
We thus explore in this work the applicability of adopting vector quantization as the latent representation in the I2I task.

We introduce VQ-I2I, a framework that adopts a vector quantized codebook as an intermediate representation which is able to enable both the image-to-image translation and the unconditional generation of input domains.
VQ-I2I consists of a joint domain-invariant content encoder, domain-specific style encoders, and domain-specific decoders.
The joint content encoder enforces a shared latent distribution among different domains.
The encoded content representation can be further decoded with the style representation obtained from the same input for realizing the self-reconstruction or with that from different inputs for achieving intra- and inter-domain translations.
Moreover, with different style representations being given, VQ-I2I is also able to perform diverse translations.

In addition to conventional image-to-image translation, we learn an auto-regressive model on the joint quantized content space to unconditionally synthesize the latent content representation.
The capability of unconditional content generation with disentangled style representation can unleash several applications:
As shown in Figure~\ref{fig:teaser}, VQ-I2I has the multifunctionality for performing I2I,  unconditional image synthesis, and image extension. 
Combining these operations, VQ-I2I can achieve extension on generated samples with the flexibility of stylizing into different domains, and image generation with transitional stylization. These cannot be done by a unified framework to the best of our knowledge.

We conduct extensive  quantitative and qualitative evaluations.
We measure the realism with the Fréchet inception distance (FID) \cite{heusel2017gans} and subjective study using the AFHQ~\cite{choi2020stargan}, Yosemite~\cite{zhu2017unpaired}, and portrait~\cite{lee2018diverse} datasets.
On the Cityscapes dataset~\cite{cordts2016cityscapes}, we use FID metric as well to compare with the I2I methods trained upon paired data.
%
%
Qualitatively, we demonstrate realistic and diverse I2I translation as well as applications including unconditional generation, image extension, completion, transitional stylization, or combinations over them. 
\section{Related Work}
\label{sec:related}

\subsection{Image-to-Image Translation.}
Image-to-image translation, first addressed in~\cite{isola2017image} , aims at learning the mapping function between the source and the target domain.
%
%
Following works focus on tackling two major challenges: how to handle unpaired data and how to model diverse translations.
Cycle-consistency is adopted to handle unpaired data~\cite{zhu2017unpaired,liu2017unsupervised}, while augmented attribute space is proposed to provide diversity~\cite{zhu2017multimodal}.
%
%
Following efforts are made to handle both challenges jointly~\cite{huang2018multimodal,lee2018diverse,lee2020drit++}, one-sided translation without cycle-consistency~\cite{park2020contrastive}, to improve the diversity~\cite{mao2019mode,mao2022continuous}, and to better handle geometric transformations~\cite{kim2019u}.
%
Take a step forward, we propose a framework that can perform not only cross-domain translations, but also enable unconditional generation and image extension using the learned representation.

\subsection{Vector Quantized Generative Models.} 
Generative models can be roughly divided into two streams: implicit and explicit density estimation methods.
Generative adversarial network, the representative of the implicit method, has been dominant due to its high-fidelity synthesized images, yet suffering from instability in training.
On the other hand, explicit methods are more tractable in training but limited to relatively blurred outputs (\eg variational autoencoder (VAE)~\cite{kingma2013auto}) or in scaling due to the pixel-level auto-regressive process (\eg PixelRNN~\cite{van2016pixel} and PixelCNN~\cite{oord2016conditional}).
Recently, vector quantization (VQ) technique has adopted explicit methods to alleviate the scaling issue with quantized latent vectors serving as latent representation~\cite{oord2017neural,razavi2019generating,han2022show,zhang2021ufc}.
VQGAN then proposes a hybrid framework to first leverage GAN technique to learn VQ codebook, then adopt transformer~\cite{esser2021taming} to train an auto-regressive model on the learned VQ indices. 
In this work, we propose adopting VQ technique in the I2I task.
\begin{figure*}[ht!]
	\centering
	\subfloat[Overall architecture of vector-quantized I2I with disentangled representations.]{%
		\includegraphics[width=\linewidth]{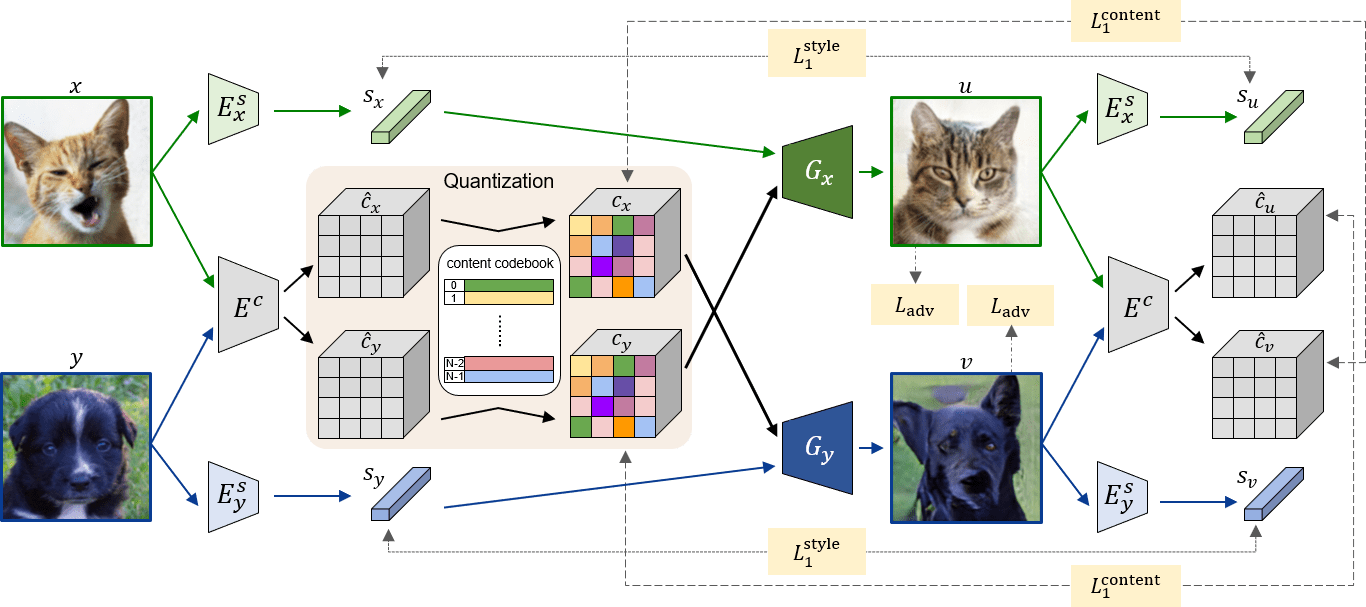}%
    }
	
    \begin{minipage}[b]{0.25\textwidth}
    \subfloat[Transformer.]{%
		\includegraphics[width=\linewidth]{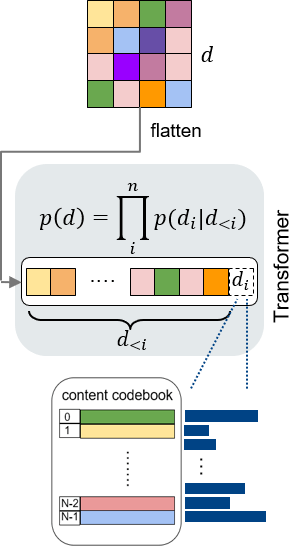}%
	}
    \end{minipage}
    \hfill
    \begin{minipage}[b]{0.71\textwidth}
    \subfloat[Image extension.]{%
		\includegraphics[width=\linewidth]{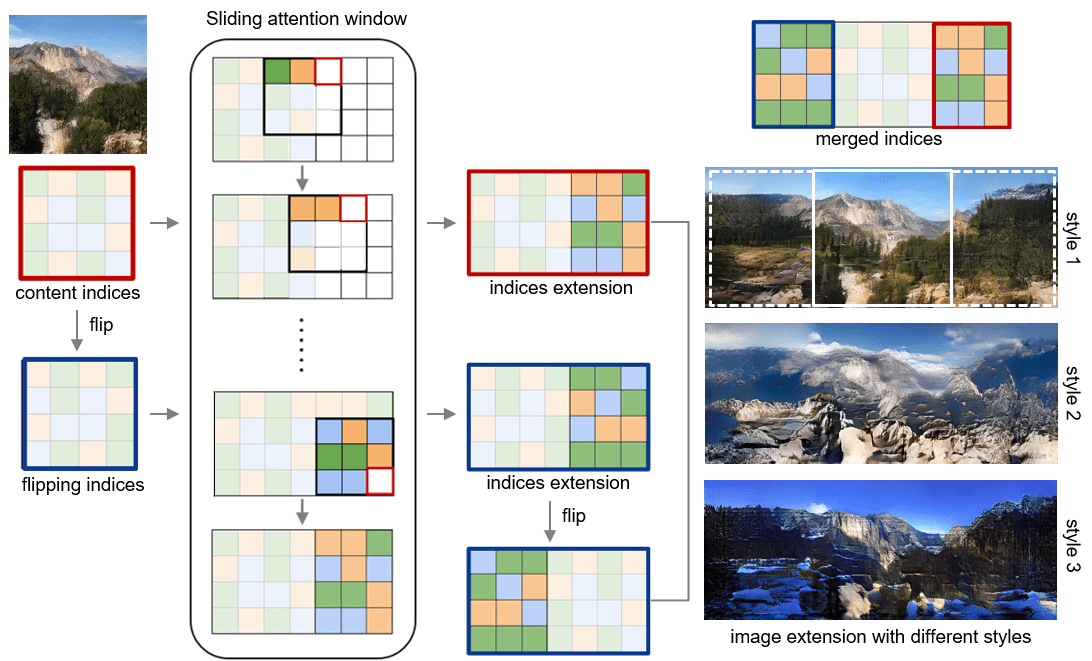}%
	}
    \end{minipage}
    \caption{\textbf{Method Overview.} (a) The proposed framework learns to perform translation with disentangled vector-quantized domain-invariant content and domain-specific style representations. (b) Given the quantized content indices $d$, we can learn the content distribution in an autoregressive manner using a transformer model. (c) With learned transformer model and the translation model, we can expand an image on both horizontal sides by spatially extending the content map and its flipped one with a sliding attention window. The extended content can be further translated into different styles.}
    \label{figure:architecture}
\end{figure*}

\label{sec:method}

\section{Method}

\indent As previously motivated, 
our goal is to leverage the vector quantized codebook, an intermediate representation for 1) image-to-image translation between two visual domains {\small $X \subset \mathbb{R}^{{H}\times{W}\times{3}}$} and {\small $Y \subset \mathbb{R}^{{H}\times{W}\times{3}}$} and 2) unconditional generation in each domain.
As illustrated in Figure~\ref{figure:architecture} (a), our framework consists of a shared content encoder $E^c$, a vector quantized content codebook $Z$, style encoders $\{{E^{s}_{X}, E^{s}_{Y}}\}$, generators $\{{G_{X}, G_{Y}}\}$, and discriminators $\{{D_{X}, D_{Y}}\}$.
Given an input image, the content encoder $E^c$ extracts the \emph{vector-quantized} domain-invariant representations, while the style encoders ${E^{s}_{X}, E^{s}_{Y}}$ compute the domain-specific features for domain $X$ and $Y$ respectively.
The generators ${G_{X}, G_{Y}}$ combine the content representation and style feature to produce the image in each domain.
Finally, the discriminators ${D_{X}, D_{Y}}$ aim to distinguish between the generated and real images.

\subsection{Vector Quantized Content Representation}\label{sec:VQ}
Our approach leverages the vector quantization strategy to encode the domain-invariant image content information.
Specifically, we construct a codebook {\small $Z = {\{z_k\}}^K_{k=1}$} that consists of learned content codes $z_k \in \mathbb{R}^{n_c}$, where $n_c$ indicates the code dimension.
Given a continuous map $\hat{c} \in \mathbb{R}^{h\times w\times n_c}$ extracted by the content encoder $E^c$, we find for each spatial entry $\hat{c}_{ij} \in \mathbb{R}^{n_c}$ of $\hat{c}$ its closest code in the codebook $Z$ for obtaining the vector quantized content representation $c$:
%
\begin{equation}
c = \mathbf{vq}(\hat{c}) := ({\arg\min}_{z_k \in Z} \|\hat{c}_{ij} - z_k\|) \in \mathbb{R}^{{h}\times{w}\times{n_c}}.
\label{eq:1}
\end{equation}

Since the quantization operation $\mathbf{vq}$ is not differentiable for gradient back-propagation, we use the straight-through trick~\cite{oord2017neural} that copies the gradient from $c$ to $\hat{c}$.
We learn the codebook $Z$ using the self-reconstruction path and the loss function $L_\mathrm{vq}$ and $L^\mathrm{recon}_1$, where
\begin{equation}
L_\mathrm{vq} = \|\mathrm{sg}[\hat{c}] - c\|^2_2 + \|\mathrm{sg}[c] - \hat{c}\|^2_2,
\label{eq:2}
\end{equation}
where 
$\mathrm{sg}[\cdot]$ is the stop-gradient operation.
We provide the details of the self-reconstruction path and the loss $L^\mathrm{recon}_1$ later in Section~\ref{sec:self-rec}.

\subsection{Diverse Image-to-Image Translation}
To enable multi-modal image-to-image translation, our approach learns the disentangled domain-\emph{invariant} content representations and domain-\emph{specific} style features~\cite{lee2018diverse,liu2017unsupervised}.
As shown in Figure~\ref{figure:architecture}(a), we use an \emph{shared} encoder $E^c$ to extract the content representation for images of two domains, followed by applying the vector quantization operation $\mathbf{vq}$, and use separate encoders ${E^{s}_{X}, E^{s}_{Y}}$ to compute the style features:
\begin{equation}
\begin{split}
&{c_x, s_x} = {\mathbf{vq}(E^c(x)), E^s_X(x)}\\
&{c_y, s_y} = {\mathbf{vq}(E^c(y)), E^s_Y(y)}.
\end{split}
\label{eq:3}
\end{equation}
Since the content space is shared among two domains, we can perform the image-to-image translation by swapping the content representations $c_x$ and $c_y$.
Finally, the generators $G_X, G_Y$ use AdaIN normalization layers~\cite{huang2017arbitrary,huang2018multimodal} to combine the swapped content representations and style features to synthesize the translated images $u \in X$ and $v \in Y$:
\begin{equation}
u = G_X(c_y, s_x),\:v = G_Y(c_x, s_y).
\label{eq:6}
\end{equation}

\subsubsection{Image-to-Image Translation Training.}
We use the discriminators $D_X$ and $D_Y$ to impose the domain adversarial loss $L_\mathrm{adv}$.
The loss $L_\mathrm{adv}$ encourages the realism of the translated images $u$ for domain $X$ and $v$ for $Y$. 
 
%
Nevertheless, training our model with the domain adversarial loss along cannot guarantee the disentanglement of content and style representations.
The content map $c$ may encode the style information, thus the generator ignores the style feature $s$ for synthesizing the translated images.
To address this issue, we use the latent style regression loss to enforce the bijection between the style features and the translated images:
\begin{equation}
\begin{split}
L^\mathrm{style}_1 &= \| E^s_X(G_X(c_y, s_x))- s_x\|\\
&+\| E^s_Y(G_Y(c_x, s_y))- s_y\|.
\end{split}
\label{eq:5}
\end{equation}
We also use the latent content regression loss to facilitate the training:
\begin{equation}
\begin{split}
L^\mathrm{content}_1 &= \| E^c(G_X(c_y, s_x))- c_y\|\\
&+\| E^c(G_Y(c_x, s_y))- c_x\|,
\end{split}
\label{eq:4}
\end{equation}
where Figure~\ref{figure:architecture} (a) shows the computation flows behind $L^\mathrm{style}_1$ and $L^\mathrm{content}_1$.

\subsubsection{Self-Reconstruction Training.}\label{sec:self-rec}
In addition to image-to-image translation, we also involve a self-reconstruction path (\ie reconstructing an image by using its own content and style representations) during the training stage for two empirical reasons.
First, self-reconstruction training is vital for learning a meaningful vector-quantized codebook~\cite{oord2017neural}.
Second, it facilitates the overall image-to-image training process.
Specifically, we impose the self-reconstruction loss:

\begin{equation}
L^{\mathrm{recon}}_1 = \|G_X(c_x, s_x)- x\|+\|G_Y(c_y, s_y)- y\|.
\label{eq:7}
\end{equation}
As described in Section~\ref{sec:VQ}, we only apply the vector quantization loss $L_\mathrm{vq}$ (cf. Equation~\ref{eq:2}) in the self-reconstruction path.
The full objective function of our model ($L_{D}$ for training discriminators; $L_{E^c, Z, E^s, G}$ for training encoders, codebook, and generators) is then summarized as:
\begin{equation}
\begin{split}
L_{D} &= L_\mathrm{adv},\\
L_{E^c, Z, E^s, G} &= -\lambda_\mathrm{adv} L_\mathrm{adv} + \lambda^\mathrm{recon}_1 L^\mathrm{recon}_1 + \lambda_\mathrm{vq} L_\mathrm{vq}\\
& + \lambda^\mathrm{content}_1 L^\mathrm{content}_1 + \lambda^\mathrm{style}_1 L^\mathrm{style}_1,
\end{split}
\label{eq:9}
\end{equation}
where $\lambda$ controls the importance of each loss term.
Note that we only optimize the codebook with the vector quantization loss $L_\mathrm{vq}$ and reconstruction loss $L^\mathrm{recon}_1$.

\subsection{Unconditional Generation}\label{sec:transformer}
Vector quantization on the shared content space enables unconditional generation, since we can model the domain-invariant joint (content) distribution using an autoregressive approach~\cite{chen2020generative}.
We present the approach in Figure~\ref{figure:architecture} (b).
The spatial entries in the content representation $c$ can be represented as a set of indices $d$ in the codebook {$\small Z = \{z_k\}^K_{k=1}$}, where $c_{ij} = z_{d_{ij}}$.
By ordering the index set $d$ using a particular rule, the content generation task can be formulated as the next-index prediction problem.
Specifically, given content indices $d_{<i}$, the goal is to predict the distribution of next index $d_i$: $p(d) = \prod_i p(d_i|d_{<i})$.
We train a transformer network~\cite{esser2021taming} for this task by maximizing the log-likelihood of the content representation: 
\begin{equation}
L_{\mathrm{transformer}} = \mathbb{E}_{x \sim p(x)}[-\mbox{log}p(d)].
\label{eq:10}
\end{equation}
%
We provide the ordering details (\ie slight difference between training and testing stages as similar to~\cite{esser2021taming}) in the supplementary materials.

During inference, we first generate the complete content representation using the autoregressive next-index prediction process.
Then we combine some style features $\{s_x, s_y\}$, and use the generators $\{G_X, G_Y\}$ to synthesize the image for different domains.

\begin{figure*}[t]
\centering
\setlength\tabcolsep{1pt} 

\subfloat[Unpaired I2I Comparison]{
    \begin{tabular}{c:ccccccc}
     \tiny Input & \tiny VQ-I2I & \tiny uni-VQ-I2I & \tiny CycleGAN & \tiny DRIT & \tiny MUNIT & \tiny U-GAT-IT & \tiny CUT \\
    \includegraphics[width=.12\linewidth]{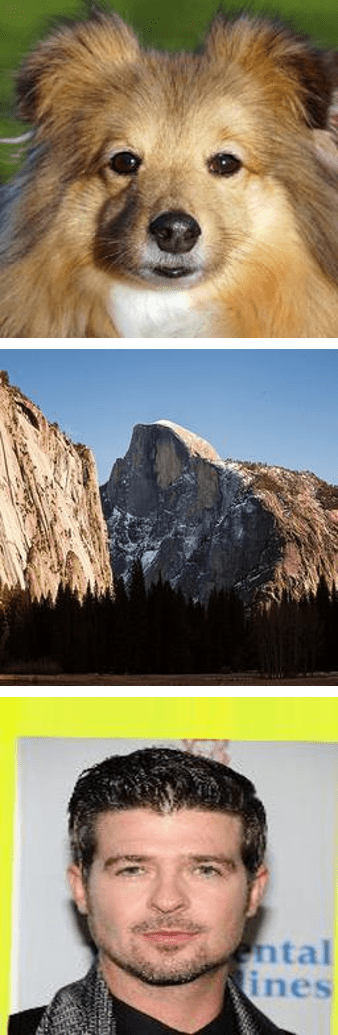} &
    \includegraphics[width=.12\linewidth]{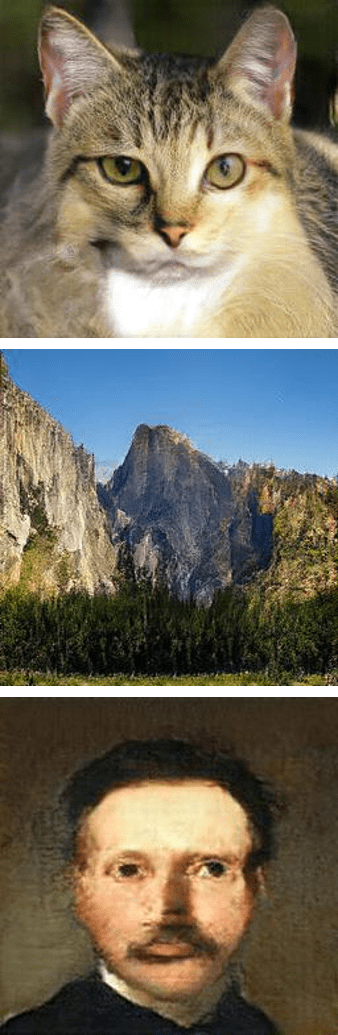} &
    \includegraphics[width=.12\linewidth]{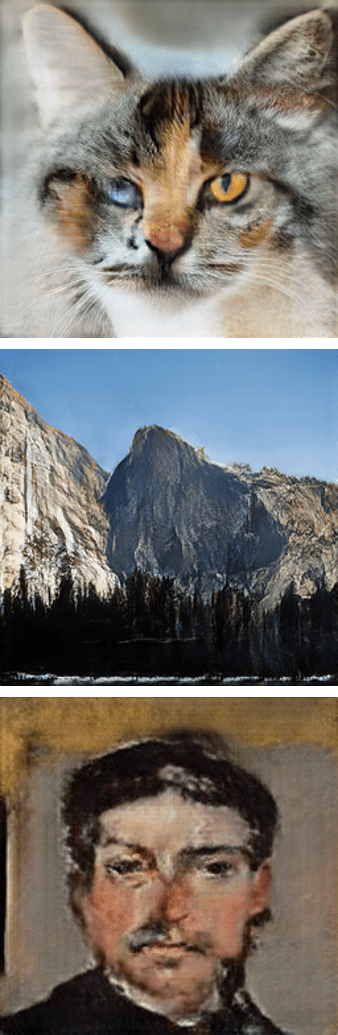} &
    \includegraphics[width=.12\linewidth]{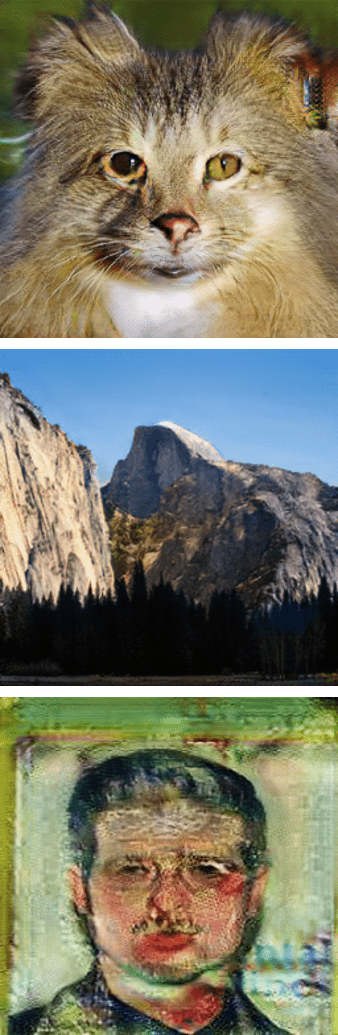} &
    \includegraphics[width=.12\linewidth]{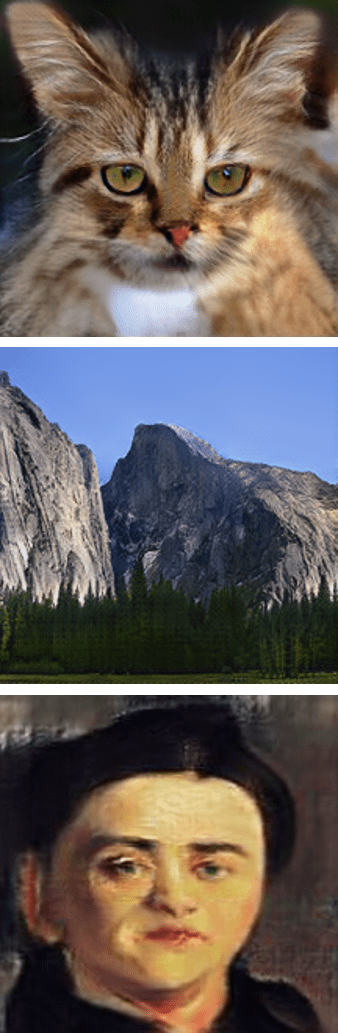} &
    \includegraphics[width=.12\linewidth]{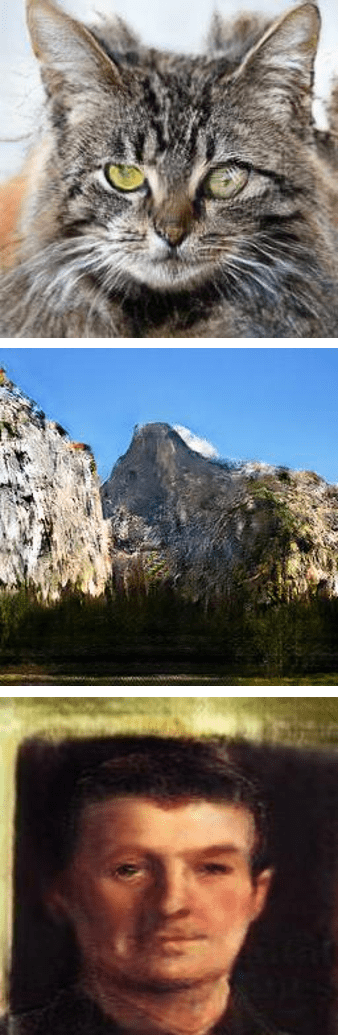} &
    \includegraphics[width=.12\linewidth]{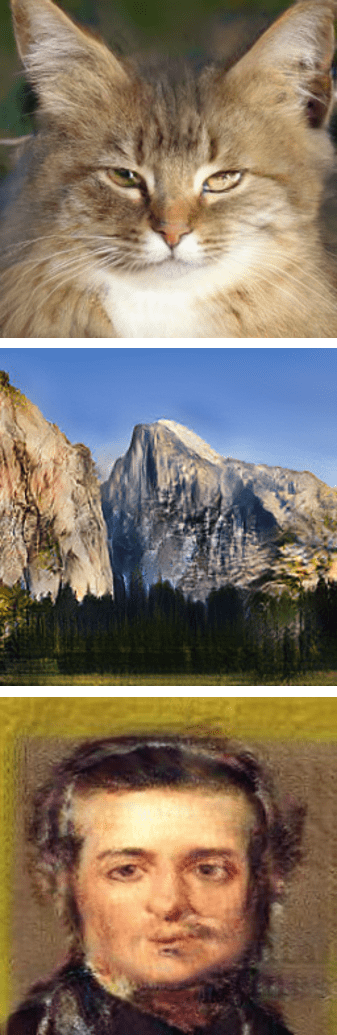} &
    \includegraphics[width=.12\linewidth]{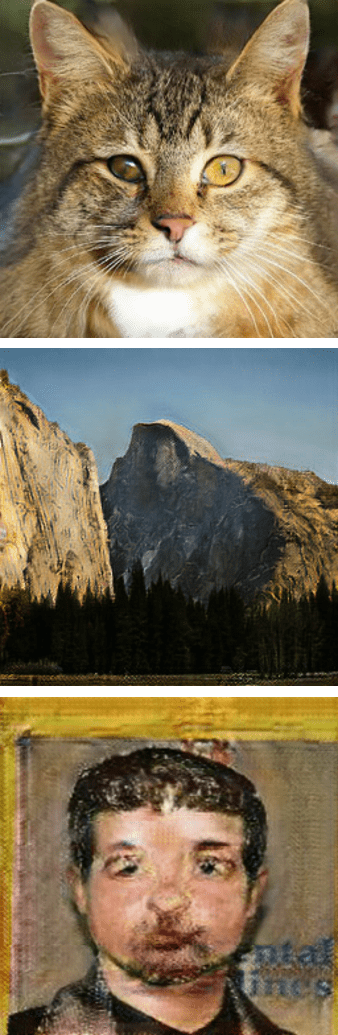} \\
    \end{tabular}
}

\subfloat[Paired I2I Comparison]{
    \begin{tabular}{c:ccc|c:ccc}
        \tiny Input & \tiny VQ-I2I & \tiny Pix2pix & \tiny BicycleGAN & \tiny Input & \tiny VQ-I2I & \tiny Pix2pix & \tiny BicycleGAN \\
    \includegraphics[width=.12\linewidth]{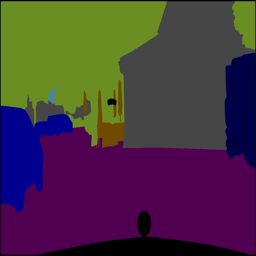} & 
    \includegraphics[width=.12\linewidth]{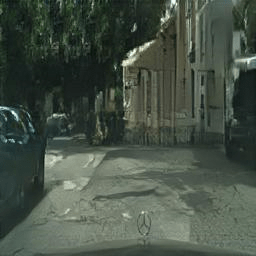} & 
    \includegraphics[width=.12\linewidth]{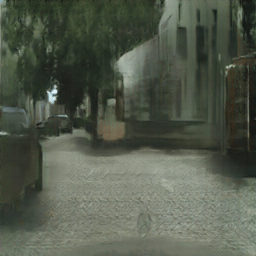} & 
    \includegraphics[width=.12\linewidth]{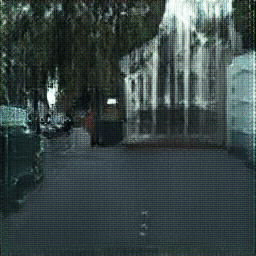} & \includegraphics[width=.12\linewidth]{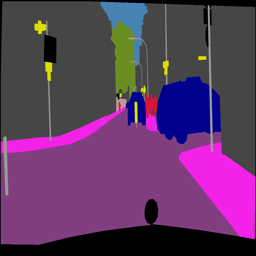} & 
    \includegraphics[width=.12\linewidth]{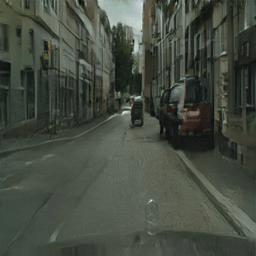} & 
    \includegraphics[width=.12\linewidth]{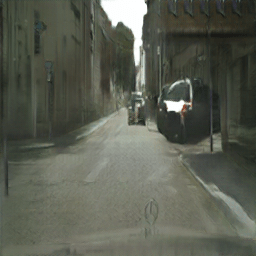} & 
    \includegraphics[width=.12\linewidth]{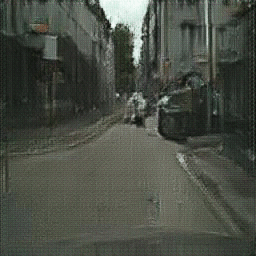} \\
    \end{tabular}
}

\caption{
\textbf{Qualitative Comparisons with Conventional Image-to-Image Translation Methods.}
(a) We show the translated results of different methods on three unpaired datasets. From top to bottom rows are dog$\rightarrow$cat~\cite{choi2020stargan}, winter$\rightarrow$summer~\cite{zhu2017unpaired}, and photo$\rightarrow$portrait~\cite{lee2018diverse}.
(b) Our model is able to handle training with paired data on Cityscapes dataset~\cite{cordts2016cityscapes}. For each example set (composed of four columns), the leftmost column shows the semantic segmentation of street scenes, and the other columns show the corresponding generated scenes by various models which are trained on paired data.
}
\label{fig:visual-unpair_and_pair}
\end{figure*}
\subsection{Content Extension.}
Our autoregressive next-index prediction process enables not only the unconditional content generation, but also \emph{content extension}: extending the content of existing images.
We illustrate the process in Figure~\ref{figure:architecture} (c).
Specifically, given a vector quantized content representation extracted from an existing image, we use the learned transformer model to spatially extend the content map (red outline).
By flipping the content representation horizontally, we can extend the content to the opposite direction using the same process (blue outline). The resultant content map which has been extended (on both horizontal sides) then can be gone through generators together with a style feature to produce the extension.




\begin{figure}[t]
\setlength\tabcolsep{1pt} 
\begin{tabular}{c:c:c|c:cc}
\tiny Input & \tiny Inter-domain & \tiny Intra-domain &
\tiny Input & \tiny Completion & \tiny +Inter-I2I \\

\includegraphics[width=.12\linewidth]{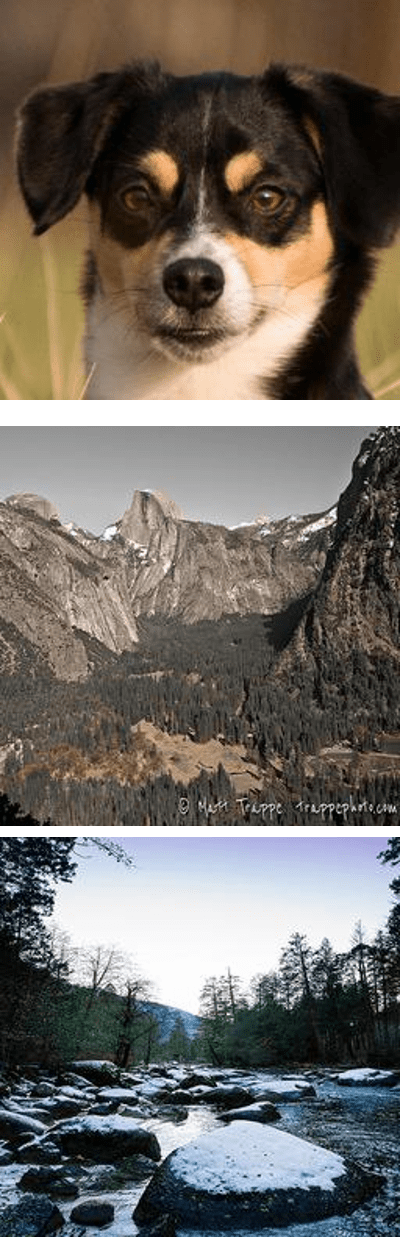} &
\includegraphics[width=.242\linewidth]{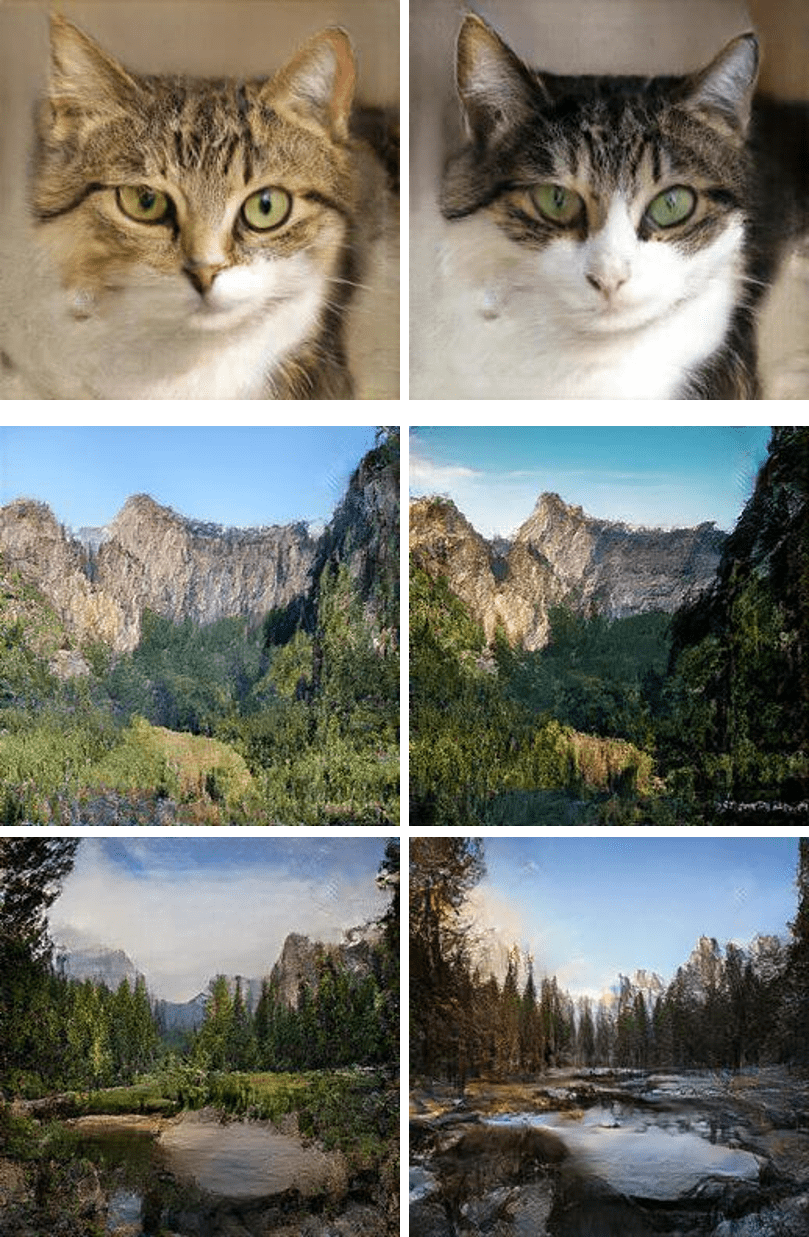} &
\includegraphics[width=.242\linewidth]{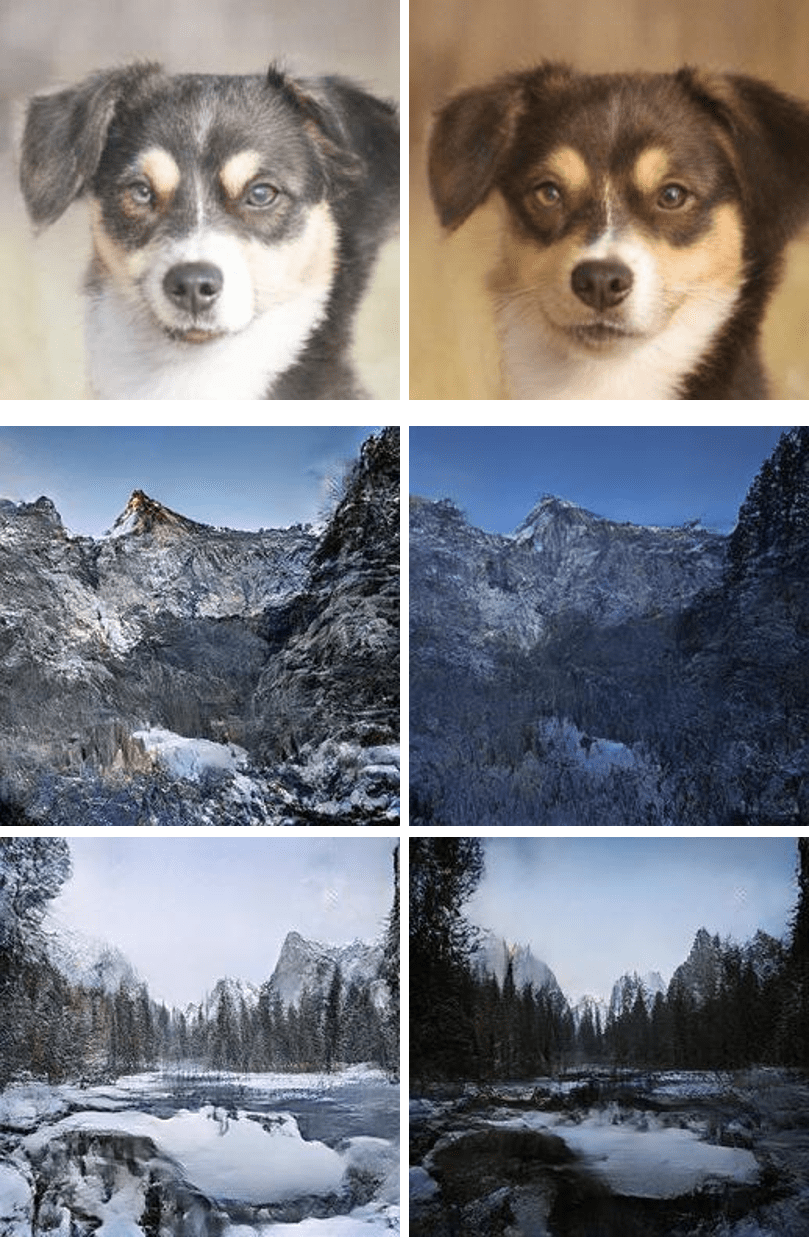} &
\includegraphics[width=.12\linewidth]{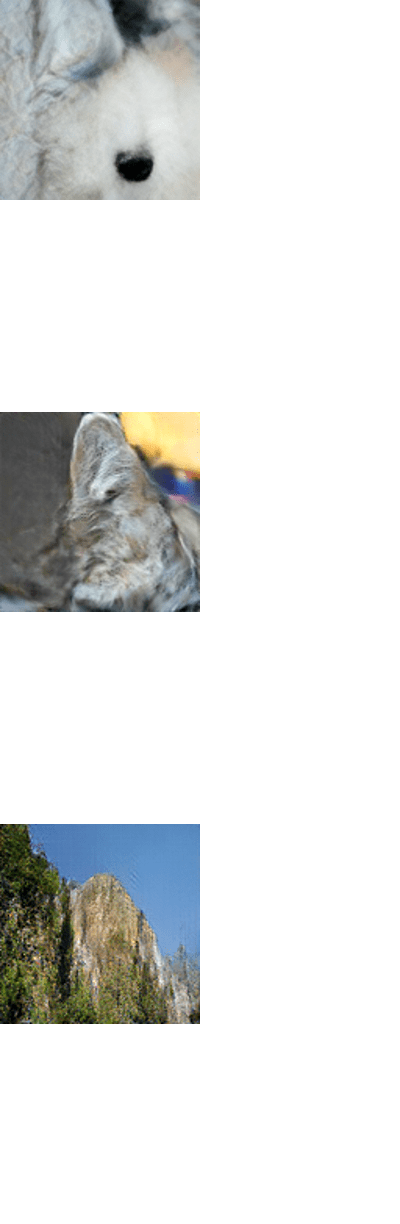} &
\includegraphics[width=.12\linewidth]{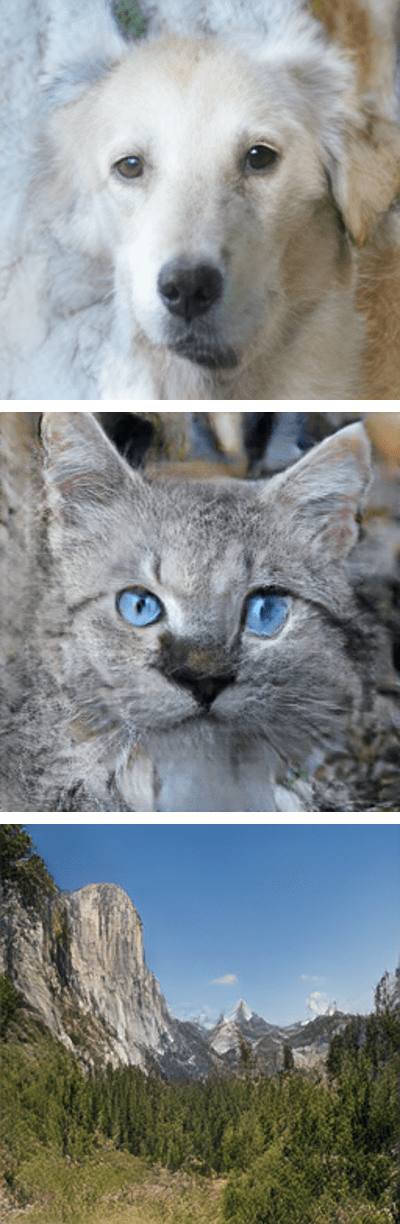} &
\includegraphics[width=.12\linewidth]{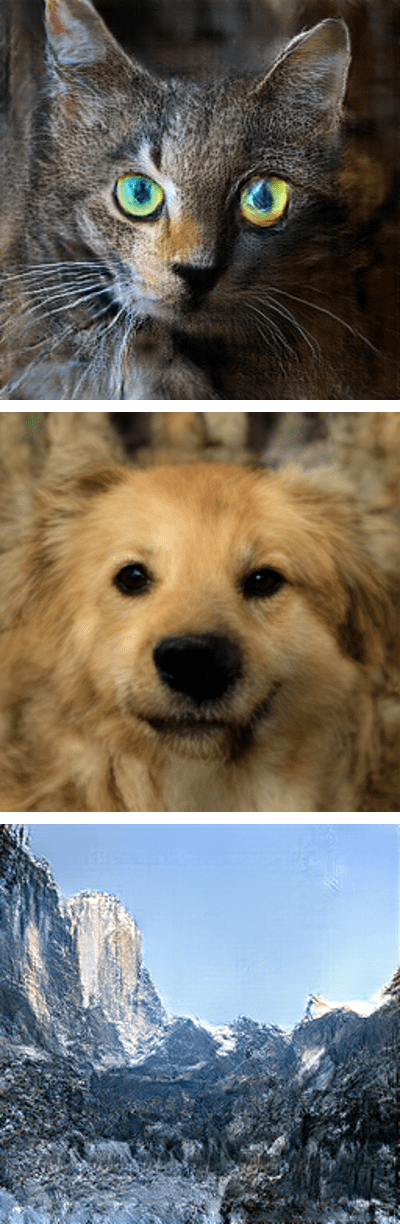}
\\

\end{tabular}
\caption{
\textbf{Diverse Image Translation and Completion.} 
(\textit{left}) We demonstrate both inter-domain and intra-domain translations with the query images (leftmost column) combined with various styles on the dog$\rightarrow$cat and winter$\rightarrow$summer scenarios. 
\textit{(right)} Given a quarter of an image from AFHQ~\cite{choi2020stargan} or Yosemite~\cite{zhu2017unpaired} dataset as the input, we perform image completion AND the inter-domain translation. VQ-I2I is able to not only learn the joint content distribution of both domains, thus achieving reasonable completion, but support the diverse translation via the design of the disentanglement.
}
\label{fig:visual-diverse}
\end{figure}
\section{Experiments}
\label{sec:experiment}
We evaluate the proposed framework on image translation, unconditional generation and image extension. We compare VQ-I2I with several representative I2I, image generation and outpainting approaches.
We then demonstrate various applications of our framework which seamlessly combine I2I with unconditional image generation, image extension, and transitional stylization.
Finally, we conduct the ablation study to understand the efficacy of different design choices.


\subsubsection{Datasets.}
We conduct experiments using both paired and unpaired I2I datasets.
For unpaired datasets, we use the Yosemite dataset~\cite{zhu2017unpaired} for the shape-invariant translation, and the AFHQ~\cite{choi2020stargan} and portrait~\cite{lee2018diverse} datasets for  the shape-variant translation task.
For paired dataset, we use the Cityscapes dataset~\cite{cordts2016cityscapes}.

\subsubsection{Compared Baselines.}
For the unpaired I2I setting, we compare our method with CycleGAN~\cite{zhu2017unpaired}, DRIT~\cite{lee2018diverse}, MUNIT~\cite{huang2018multimodal}, and recent CUT~\cite{park2020contrastive} and U-GAT-IT~\cite{kim2019u}.
For the paired I2I setting, we make a comparison between our method and Pix2pix~\cite{isola2017image} as well as BicycleGAN~\cite{zhu2017toward}.
For unconditional generation, we compare our approach with VQGAN~\cite{esser2021taming}.
As for image extension~\cite{cheng2022inout,lin2021infinitygan,teterwak2019boundless}, we consider a representative baseline from Boundless~\cite{teterwak2019boundless}.
The training details are provided in the supplement.\par

Furthermore, to understand the impact of having the latent representation explicitly disentangled, we construct an uni-modal VQ-I2I variant as an additional baseline (denoted as uni-VQ-I2I).
Specifically, in such uni-VQ-I2I baseline, we assume that the domain-specific style information is implicitly modeled by the generators $\{G_X, G_Y\}$, thus the domain-specific style features are discarded.
Please refer to our supplementary materials for more details.

\begin{figure*}[t]
\centering
\setlength\tabcolsep{1.5pt} 
\includegraphics[width=.99\linewidth]{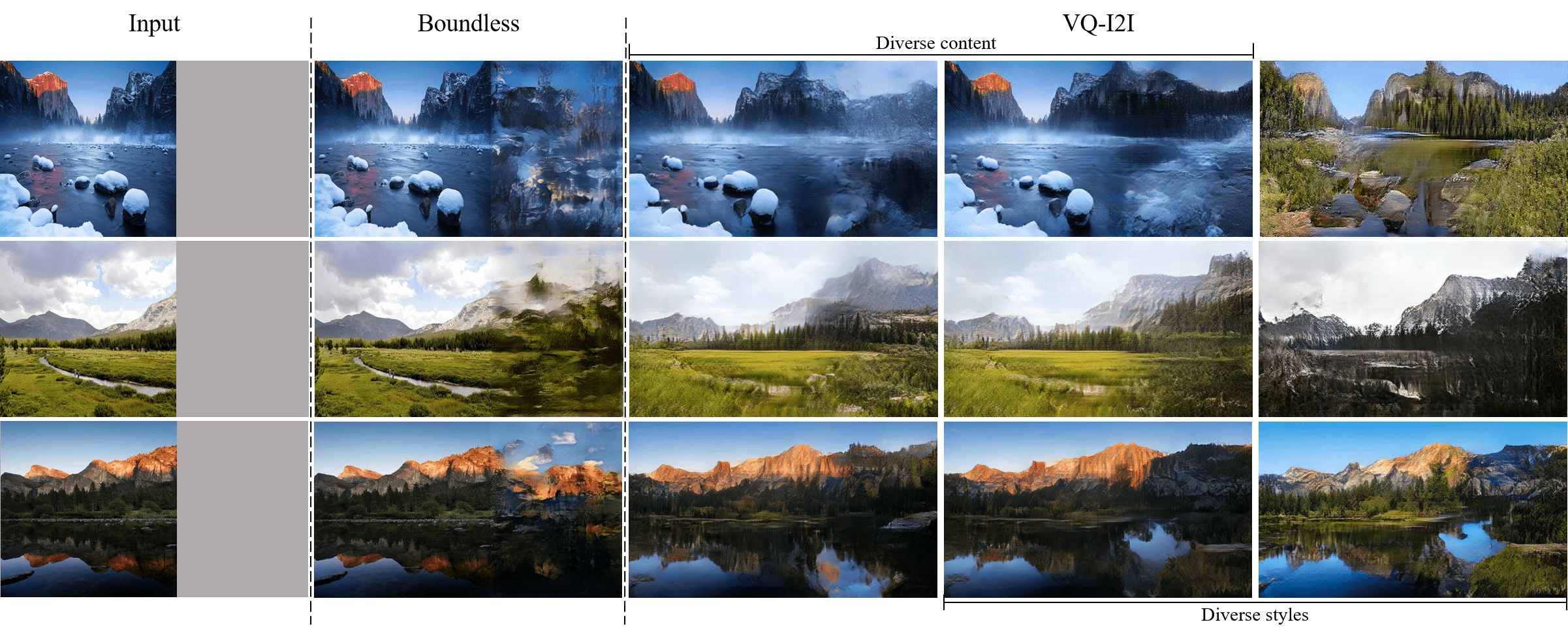} \\
\caption{\textbf{Qualitative Examples on Image Extension.} 
Example results of image extension on Yosemite~\cite{zhu2017unpaired} datasets, where the comparison with respect to Boundless~\cite{teterwak2019boundless} baseline is also provided.
The leftmost column shows the input images for the image extension, where the model takes left portion of size 256$\times$256 for each input image and extends for 192 pixel width toward the right-hand side. VQ-I2I is able to generate smooth and diverse extensions with style variability.  
}
\label{fig:yosemite-outpaint}
\end{figure*}
\subsection{Qualitative Evaluation}
\subsubsection{I2I Translation on Unpaired and Paired Data.} 
The proposed VQ-I2I synthesizes high-quality images on both the shape-invariant (winter-to-summer) and shape-variant (dog-to-cat, photo-to-portrait) datasets, as shown in Figure~\ref{fig:visual-unpair_and_pair}(a), where it achieves comparable or even better quality in comparison to other representative I2I methods.
Moreover, the results of uni-modal VQ-I2I variant (denoted as uni-VQ-I2I), which excludes the disentanglement between content and style information are also provided, where we are able to observe that uni-VQ-I2I encounters the problem of texture inconsistency (\eg there exists different styles in the cat's face on the first row of Figure~\ref{fig:visual-unpair_and_pair}(a)).
The comparison between our VQ-I2I and the uni-VQ-I2I variant reveals that the characteristic of disentanglement enables both content stability and style diversity, where we provide further explorations on uni-VQ-I2I in the supplementary materials.
%
%
On the other hand, given pairs of semantic segmentation maps and corresponding images as training data, our proposed scheme produces appealing images that correspond to the input segmentation map (cf. Figure~\ref{fig:visual-unpair_and_pair}(b)).
These results validate that our VQ-I2I approach can understand the semantic meaning of labels and synthesize correct instances, such as buildings and vehicles.\par
%
\subsubsection{Multimodal Translation.}
Our VQ-I2I framework can also perform style-guided translation that produces diverse (multimodal) I2I results.
Since vector-quantized content representation encodes the domain-invariant information while the style features carry the style information, we re-combine the same content with various styles to achieve diverse translations.
The results are shown in the left portion of Figure~\ref{fig:visual-diverse}.
In addition to the inter-domain I2I, our method can also perform \emph{intra-}domain I2I (as shown in the column labelled as ``intra-domain'' of Figure~\ref{fig:visual-diverse}, in which we combine the content and style extracted from two images of the same domain), although we do not explicitly involve intra-domain I2I during the training stage.
%
%

\subsubsection{Diverse Image Extension and Completion.} 
The auto-regressive procedure built upon the content representation of VQ-I2I enables image extension. Specifically, as the content indices on the extended regions are drawn from the conditional distribution predicted by the transformer model, together with the style features being disentangled from content, the resultant extension produced by our VQ-I2I includes the diversity of both content and style (cf. Figure~\ref{fig:yosemite-outpaint}). 
%
%
It is worth noting that the extension results show that VQ-I2I generators would adjust the original image slightly to make the overall appearance of image extension more harmonious.
Similar to image extension, our VQ-I2I is able to realize the image completion. We conduct the experiments of image completion on AFHQ~\cite{choi2020stargan} and Yosemite~\cite{zhu2017unpaired} dataset and provide some example results in the right portion of Figure~\ref{fig:visual-diverse}, where only a quarter of an image is given as the input. Again, our auto-regressive model and the disentanglement designs are capable of generating diverse content and supporting style variability via combining the translation (\eg inter-domain I2I in the rightmost column of Figure~\ref{fig:visual-diverse}).
%
%

\begin{table}[t]
	\caption{
	 \textbf{Quantitative Comparisons with Unpaired I2I Methods.} We measure the FID and NIQE scores across various datasets. VQ-I2I performs comparably to the state-of-the-art methods on unpaired datasets, while enabling applications that cannot be done by these conventional I2I methods.
	}
	\label{tab:fid}
	\centering
	\scriptsize
	\tabcolsep=0.15cm
        \begin{tabular}{lccccc} 
        \toprule
         & \multicolumn{3}{c}{\textbf{FID}} & \multicolumn{2}{c}{\textbf{NIQE}} \\ 
        \cmidrule(lr){2-4}\cmidrule(l){5-6}
         & dog$\rightarrow$cat & winter$\rightarrow$summer & photo$\rightarrow$portrait & dog$\rightarrow$cat & winter$\rightarrow$summer \\ 
        \midrule
        CycleGAN~ & 76.89 & 65.71 & 104.96 & \textbf{40.65} & 53.28 \\
        DRIT~ & 35.74 & \textbf{60.53} & 102.52 & 47.32 & \textbf{32.76} \\
        MUNIT~ & 33.78 & 94.78 & \textbf{94.42} & 63.64 & 35.64 \\
        U-GAT-IT~ & \textbf{21.62} & 73.89 & 104.93 & 59.84 & 57.44 \\
        CUT~ & 22.79 & 70.41 & 102.65 & 48.95 & 37.29 \\ 
        \hdashline
        uni-VQ-I2I & 25.65 & 62.43 & 99.37 & 45.41 & 36.53 \\
        \textbf{VQ-I2I} & 26.53 & 63.60 & 100.29 & 53.29 & 35.97 \\
        \bottomrule
        \end{tabular}
        \label{tab:unpair-fid}
\end{table}
\begin{table}[t!]
\centering
\small
\caption{
{
\textbf{Quantitative Comparisons on Applications of VQ-I2I.} (a) We evaluate the performance (in terms of FID scores) of unconditional generation on Yosemite~\cite{zhu2017unpaired} dataset via sampling 100 images respectively generated by our VQ-I2I and the VQGAN~\cite{esser2021taming}. (b) Given the input image of size 256$\times$256, we extend it horizontally for 50\% and 75\% (128 and 192 pixels respectively) toward the right-hand size, where we evaluate the FID scores on the right most portion of size 256$\times$256 of the resultant image (\ie this portion will recover part of the original input image and the extended region).
The results show that our model is comparable to the existing extension method while extending for a larger range.
}
}
    \begin{minipage}[b]{0.42\textwidth}
    \subfloat[Unconditional Generation.]{%
        \centering
    	\tabcolsep=0.2cm
        \scriptsize
		\begin{tabular}{lc}
        \toprule
         & 256$\times$256 generation\\ 
        \midrule
        VQGAN & 127.84 \\
        \textbf{VQ-I2I} & 127.31 \\
        \bottomrule
        \end{tabular}
        \label{tab:uncond_gen}
	}
    \end{minipage}
    \hspace{-3mm}
    \begin{minipage}[b]{0.53\textwidth}
    \subfloat[Image Extension.]{%
    	\tabcolsep=0.15cm
    	\scriptsize
        \begin{tabular}{lcc}
        \toprule
         & outpaint for 50\% & outpaint for 75\% \\ 
        \midrule
        Boundless~\cite{teterwak2019boundless} & \textbf{68.00} & \textbf{88.95} \\
        \textbf{VQ-I2I} & 77.82 & 90.05 \\
        \bottomrule
        \end{tabular}
        \label{tab:outpaint_fid}		
	}
    \end{minipage}
    \label{tab:application-fid}
\end{table}

\begin{figure}[ht]
	\centering
    \includegraphics[width=\linewidth]{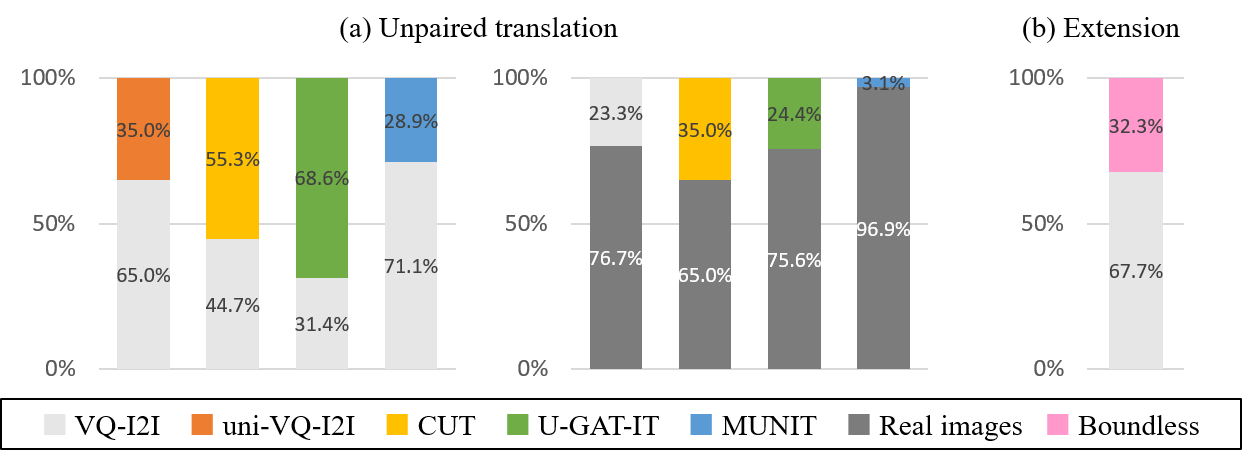}
    \caption{
    \textbf{User Preference Study.} We conduct the user study ($\sim$180 participants) to compare VQ-I2I to different existing translation methods in (a), and the Boundless~\cite{teterwak2019boundless} method for the image extension task in (b).
    }
    \label{fig:user-study}
\end{figure}

\begin{table}[t]
\tabcolsep=0.1cm
\begin{minipage}[b]{0.47\textwidth}
    \caption{
	 \textbf{Quantitative Comparisons with Paired I2I Methods.} We measure FID score for label$\rightarrow$cityscapes translation on Cityscapes~\cite{cordts2016cityscapes} Dataset.}
	\scriptsize
	
    \tabcolsep=0.2cm
	\begin{center}
	    
	\begin{tabular}{lc} 
    \toprule
     & \textbf{FID} \\
    \cline{2-2}
     & label→cityscapes \\
    \midrule
    Pix2pix~\cite{isola2017image} & \textbf{51.73} \\
    BicycleGAN~\cite{zhu2017toward} & 93.13 \\
    \hdashline
    \textbf{VQ-I2I} & 74.03 \\
    \bottomrule
    \end{tabular}
    \end{center}
    \label{tab:pair-fid}
\end{minipage}
\hspace{2mm}
\begin{minipage}[b]{0.47\textwidth}
    \caption{\textbf{Ablation of Varying the Codebook Size and Dimensionality.} We measure the FID score for summer$\rightarrow$winter translation after 420 epochs of training on each model.}
	\scriptsize
	\tabcolsep=0.15cm
	\begin{center}
    \begin{tabular}{lcc} 
    \toprule
     & \multicolumn{2}{c}{\textbf{Codebook Size}} \\ 
    \cline{2-3}
     & 64 & 512 \\ 
    \midrule
    Dimensionality 64 & 96.94 & 96.71 \\
    Dimensionality 512 & 94.38 & 99.51 \\
    \bottomrule
    \end{tabular}
    \end{center}
    \label{tab:ablation-codebook}
    
\end{minipage}

\end{table}
\subsection{Quantitative Evaluation}
We use the Fréchet inception distance (FID)~\cite{heusel2017gans} score and natural image quality evaluator (NIQE)~\cite{mittal2012making} to measure the quality of the generated results and compare our proposed method to the existing approaches. Lower FID and NIQE scores indicate better perceptual quality. 
Moreover, we conduct a user study using the manner of pairwise comparison (\ie our VQ-I2I versus baselines, or the images produced by various methods against the real images).
 
%
\subsubsection{FID and NIQE.}
We summarize the FID and NIQE evaluation of unpaired I2I translation in Table~\ref{tab:fid} and FID measurement of paired I2I translation in Table~\ref{tab:pair-fid}. For unconditional generation, we compute the FID scores on the synthesized image of size 256$\times$256, as shown in Table~\ref{tab:application-fid}(a). As for image outpainting/extension, we present the quantitative results in Table~\ref{tab:application-fid}(b), where the FID scores for image extension are computed from the distribution distance between the Yosemite dataset~\cite{zhu2017unpaired} and the rightmost 256$\times$256 pixels of the extended images.\par
%
We are able to see that our proposed method performs comparably against the state-of-the-art translation frameworks, generative approach (\ie VQGAN~\cite{esser2021taming}) and the extension baselines (\ie Boundless~\cite{teterwak2019boundless} and InfinityGAN~\cite{lin2021infinitygan}).
Please note that, the main goal of our VQ-I2I is not to achieve superior performance in translation, unconditional generation or extension, instead we aim to facilitate both translation and the unconditional distribution shared among domains in a unified novel framework as well as unleash various interesting applications which other existing works are hard to realize (as described in the next subsection).\par
%
\subsubsection{User Preference.}
To better rate the realism of I2I translation and image extension results, we conduct a user study with the manner of pairwise comparison.
For I2I translation, each subject (in total $\sim$180 participants) needs to answer the question ``Which image is more realistic'' given a pair of images (1) sampled from real images and the translated images generated from various I2I baselines or (2) respectively produced by our VQ-I2I and one of the baselines; while for extension, the comparison is conducted between our VQ-I2I and a baseline extension method (\ie Boundless~\cite{teterwak2019boundless}).
Figure~\ref{fig:user-study} presents the results of the user study.
The performance of VQ-I2I is comparable to those SOTA methods in I2I translation and image extension.
\par

\subsection{Applications}\label{sec:applications}
\subsubsection{Unconditional Image Generation and Image Extension.}
VQ-I2I completes more applications that other existing pixel-level I2I models scarcely achieve, as we adapt the vector quantized representation to the disentangled domain-invariant content space. 
Combining generated or extended content codes with a replaceable style representation, VQ-I2I can be further utilized in two applications: unconditional image generation and image extension with flexible style modulation in different ways (i.e. the combination between generation/extension and intra- or inter-domain I2I), where we have demonstrate example results in Figure~\ref{fig:visual-diverse}, Figure~\ref{fig:yosemite-outpaint}, and Figure~\ref{fig:teaser} (b)(c)(d).
These applications afford to make image synthesis style-oriented, and there are more results provided in the supplementary materials.

\begin{figure*}[t]
\setlength\tabcolsep{1.5pt} 
\includegraphics[width=\linewidth]{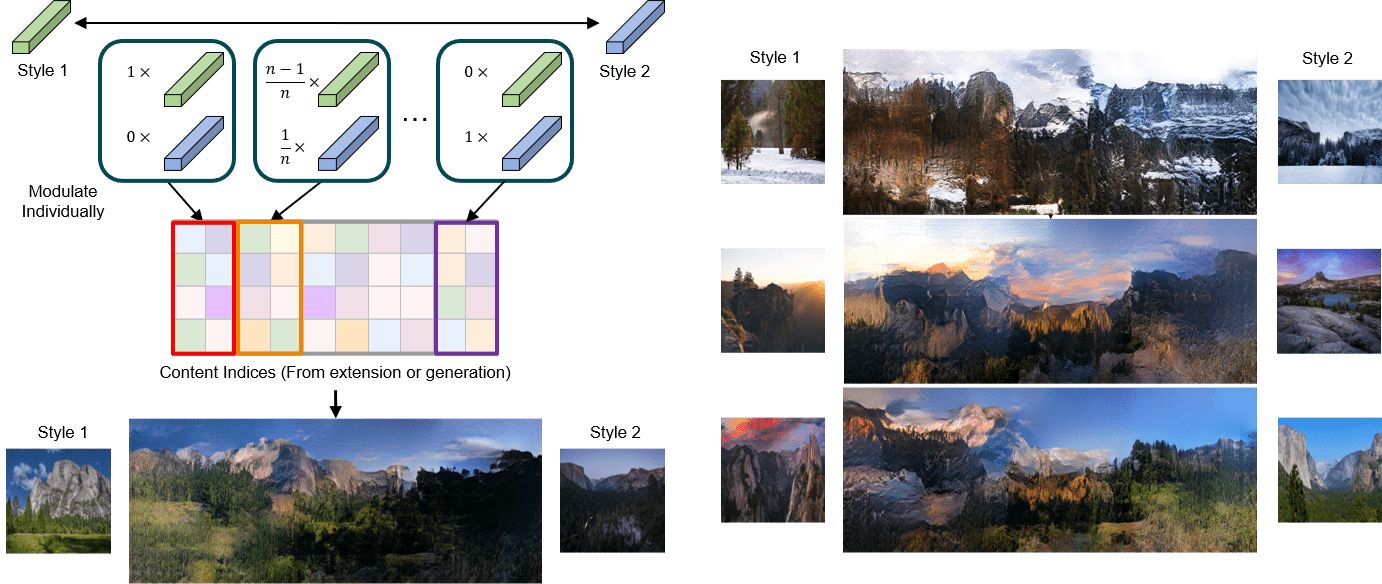} \\
\caption{
\textbf{Advanced Application of our VQ-I2I: Transitional Stylized Image Synthesis.} 
Given two guided styles and the content map (produced from extension or unconditional generation), VQ-I2I is capable of synthesizing images with a smooth and gradually changing stylization effect via blending over two styles.}
\label{fig:transitional-arch}
\end{figure*}

\subsubsection{Stylized Transitional Generation.}
In addition to single style modulation on the generated content, we can also perform multi-style transitional transfer via interpolating two styles to produce the style representation.
As shown in Figure~\ref{fig:transitional-arch}, We modulate different parts of the content map independently with different proportions by mixing the two styles, and merge all these modulated latents together to generate the transitional stylized output.
In detail, for producing smooth and gradually changing effect of stylization,
we partition the content map horizontally to 10 equal splits, where some example results are demonstrated in Figure~\ref{fig:teaser}(e) and Figure~\ref{fig:transitional-arch}.
More results with different number of splits are provided in the supplementary materials.

\begin{figure*}[t!]
    \begin{minipage}[b]{0.48\textwidth}
    \subfloat[Ablation study on adopting PatchNCE Loss in our VQ-I2I model (performance in terms of FID scores).]{%
        \small
    	\tabcolsep=0.1cm
        \begin{tabular}{lcc} 
        \toprule
        FID & VQ-I2I & +PatchNCE \\ 
        \midrule
        dog$\rightarrow$cat & \textbf{29.07} & 80.72 \\
        winter$\rightarrow$summer & \textbf{65.64} & 71.17 \\
        photo$\rightarrow$portrait & \multicolumn{1}{l}{125.37} & \textbf{114.04} \\
        \bottomrule
        \end{tabular}
        \label{tab:patchnce-fid}
	}
    \end{minipage}
    \begin{minipage}[b]{0.48\textwidth}
    \subfloat[Qualitative examples of Patchnce loss.]{
        
\begin{tabular}{cc:cc}
\tabcolsep=1pt
\tiny content & \tiny style & \tiny VQ-I2I & \tiny +PatchNCE\\
    \includegraphics[width=.24\linewidth]{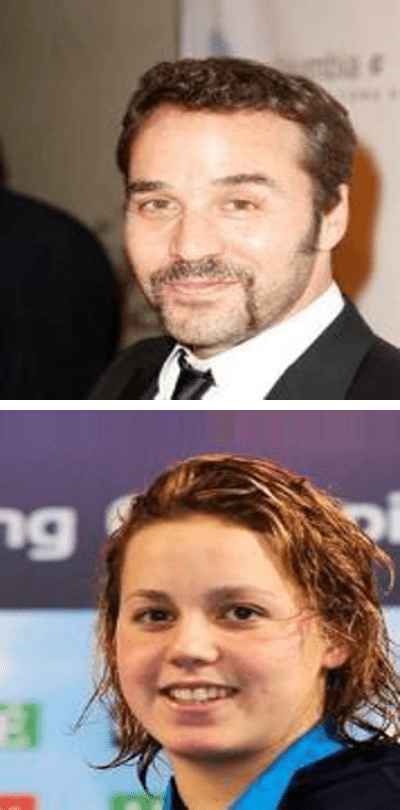} & 
    \includegraphics[width=.24\linewidth]{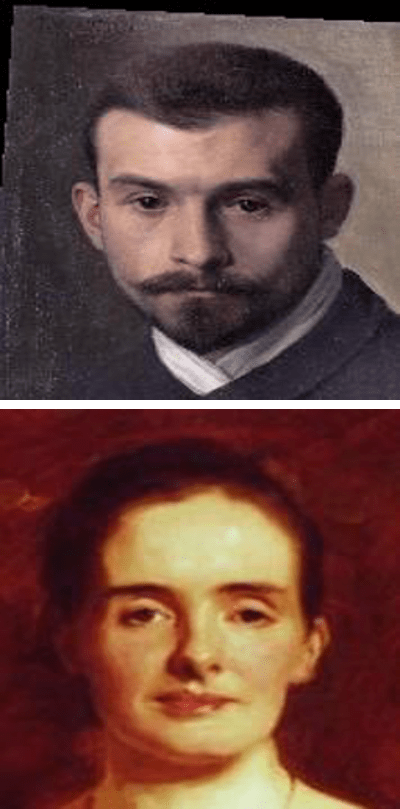} & 
    \includegraphics[width=.24\linewidth]{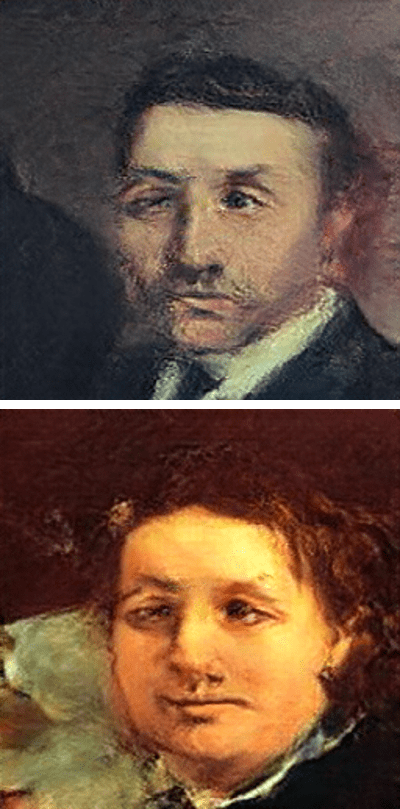} & 
    \includegraphics[width=.24\linewidth]{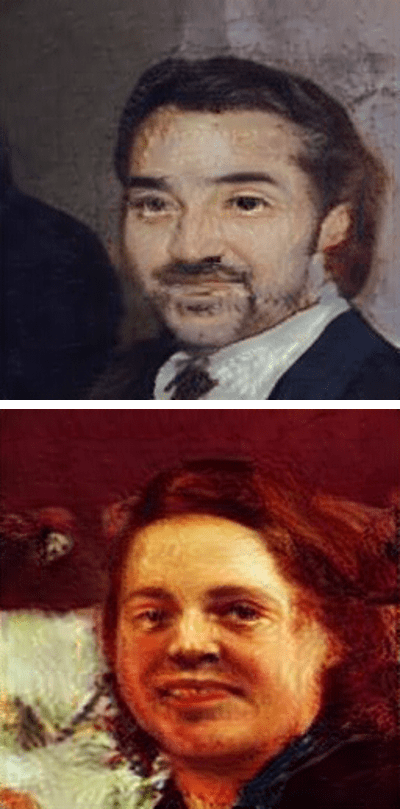} \\
    \end{tabular}
	}
    \end{minipage}
    \caption{
    \textbf{Ablation of Adding Patchwise Contrastive Loss (PatchNCE Loss).} (a) We compute the FID scores with the same input content and style images on AFHQ, Yosemite, and Portrait datasets. The quantitative results reveal that PatchNCE loss makes a strong improvement on photo$\rightarrow$portrait translation. (b) Given the input content and style images, the visual results show that PatchNCE loss is beneficial for our VQ-I2I model for preserving the content information on Portrait dataset~\cite{lee2018diverse}.}
\label{fig:patchnce}
\end{figure*}

\subsection{Further Investigation}
\subsubsection{Adding Patchwise Contrastive Loss.}
As our VQ-I2I framework does not include the cycle consistency as used in CycleGAN~\cite{zhu2017unpaired} or DRIT~\cite{lee2018diverse}, there could exist a potential concern about being unable to well preserve the geometric information during I2I translation. 
To address this issue, here we experiment to adopt the patchwise contrastive loss~\cite{park2020contrastive}, also named as PatchNCE loss, to enhance the content preservation during the training phase. As shown in Figure~\ref{fig:patchnce}(a), the performance of using PatchNCE loss is more task-sensitive. Therefore, we consider it as an optional design choice, and use the content/style regression loss as the default design in our framework.\par
\subsubsection{Varying Codebook Size and Dimensionality.}
To observe the usage of codes in the codebook, we conduct additional ablation on Yosemite dataset~\cite{zhu2017unpaired} by varying the codebook size and the dimensionality of the codebook in VQ-I2I, and the FID scores for summer$\rightarrow$winter translation is shown in Table~\ref{tab:ablation-codebook}.
When setting the codebook size as 512 and dimensionality of the codebook as 512, our VQ-I2I model only uses around 35 codes.
Besides, when shrinking both codebook size and dimensionality to 64, the codebook utilization grows up to 100\%.
However, from Table~\ref{tab:ablation-codebook}, the quantitative differences between different codebook size and dimensionality are imperceptible.
Therefore, we suggest that training on Yosemite dataset~\cite{zhu2017unpaired} for a smaller codebook size and dimensionality still maintains its performance and reduces the memory usage of our VQ-I2I model. 
\par
\section{Conclusion}
\label{sec:conclusion}
In this paper, we introduce VQ-I2I, a novel image-to-image translation framework equipped with disentangled and discrete representations. In particular, our method learns a vector-quantized codebook for capturing the domain-invariant content information of input domains, in which such codebook enables the learning of the content distribution via an autoregressive model built upon the transformer network. Upon having comparable quantitative and qualitative performance at image-to-image translation with respect to several baselines, VQ-I2I is especially novel to have multifunctionality integrated into a unified framework, including image-to-image translation, unconditional generation, image extension, transitional stylization, and the combinations of the applications above. 

\noindent\textbf{Acknowledgement.}
This project is supported by MediaTek Inc., MOST (Ministry of Science and Technology, Taiwan) 111-2636-E-A49-003 and 111-2628-E-A49-018-MY4. We are grateful to the National Center for High-performance Computing for computer time and facilities.

\clearpage
%

%
\bibliographystyle{splncs04}
\bibliography{egbib}
\end{document}


\pagestyle{headings}
\mainmatter
\def\ECCVSubNumber{4584}  

\title{Vector Quantized Image-to-Image Translation \textit{Supplementary Materials}} 

\titlerunning{Vector Quantized Image-to-Image Translation}
%
\author{Yu-Jie Chen\protect\footnotemark[1]\inst{1,2}, Shin-I Cheng\protect\footnotemark[1]\inst{1,2} 
\and
Wei-Chen Chiu\inst{1,2}\orcidlink{0000-0001-7715-8306} \and\\
Hung-Yu Tseng\inst{3} \and
Hsin-Ying Lee\inst{4}
}
\authorrunning{Y. Chen, S. Cheng et al.}
%
\institute{
$^1$National Chiao Tung University, Taiwan 
$^2$MediaTek-NCTU Research Center
$^3$Meta
$^4$Snap Inc.
}

\maketitle

\section{Dataset Details}
We consider three unpaired datasets: AFHQ dataset~\cite{choi2020stargan}, Yosemite dataset~\cite{zhu2017unpaired} and Portrait dataset~\cite{lee2018diverse}, and one paired dataset Cityscapes~\cite{cordts2016cityscapes}. AFHQ dataset has three domains: 5153 training and 500 testing images of ``cat'', 4739 training and 500 testing images of ``dog'', and 4738 training and 500 testing images of ``wildlife'' (e.g. tiger, lion, wolf, etc). Yosemite dataset contains landscape photos collected from Yosemite National Park with two classes, which are related to photos of two seasons: 1231 training and 309 testing images of ``summer'' and 962 training and 238 testing images of ``winter''. Portrait dataset has two classes (i.e., portraits in photography images and the ones in painting images),
It contains 1711 training and 100 testing images of painting portraits, and 6352 training and 100 testing images of photography portraits. Cityscapes dataset is pairwise, having 2975 training and 500 testing images of the cityscape and their corresponding semantic segmentation maps. During the training phase, we resize all images to the resolution of $256\times256$.

\section{Implementation Details}\label{sec:appendix}
We implement the models with Pytorch. The training of our whole proposed framework is divided into two stages. Firstly, we train our VQ-I2I architecture by using the objective function summarized in Equation 8 of our main manuscript to address the diverse  image-to-image translation task and build a representative vector-quantized codebook for the content information, where the details are provided later in Section~\ref{sec:VQ-I2I}. Then, in the second stage, we go on training an autoregresssive transformer model based on the content codebook and the content encoder $E^c$ learnt in the first stage to further address two applications: unconditional image generation and image extension. The details of the second stage are provided later in Section~\ref{sec:transformer-detail} and~\ref{sec:applications}. Moreover, we provide the training details of the unimodal VQ-I2I (denoted as uni-VQ-I2I) baseline in Section~\ref{sec:Uni-VQ-I2I}.

\subsection{VQ-I2I}\label{sec:VQ-I2I}

\paragraph{Settings of hyper-parameters.}
We use the Adam optimizer~\cite{esser2021taming} for model training with a batch size of 1, a learning rate of 0.00001, and exponential decay rates $(\beta_1, \beta_2) = (0.5, 0.999)$. In all experiments, we set the hyper-parameters $\lambda$ to balance between different objective functions as follows: $\lambda_\mathrm{adv} = 0.1$, $\lambda^\mathrm{recon}_1 = 5$, $\lambda_\mathrm{vq} = 1$, $\lambda^\mathrm{content}_1 = 0.2$ and $\lambda^\mathrm{style}_1 = 1$. For the vector-quantized content codebook, different parameters are set up for each of the datasets used in our experiments, as we assume that the dataset with more multifarious content needs the larger codebook size. Yosemite dataset contains rich and complex landscapes, so we adjust its content codebook with having the number of embedding set to 512 and the embedding dimensionality set to 512. For both AFHQ and Photo2Portrait datasets, the number of embeddings is set to 256 and the embedding dimensionality is set to 256. For Cityscapes dataset, we set the number of embeddings to 64 and the embedding dimensionality to 256 for the content codebook.
The settings of the content codebook for various datasets are summarized in Table~\ref{tab:codebook-setting}.

\begin{table}[ht!]
\centering
\caption{\textbf{The settings of content codebook for different datasets used in our experiments}, including the codebook size (i.e. the number of codes in a codebook) and the dimensionality of each code/embedding.}
\begin{tabular}{lcc} 
\toprule
Datasets   & codebook size & code dimensionality  \\ 
\midrule
Yosemite   & 512                  & 512               \\
AFHQ       & 256                  & 256               \\
Portrait   & 256                  & 256               \\
Cityscapes & 64                   & 256               \\
\bottomrule
\end{tabular}
\label{tab:codebook-setting}
\end{table}

\paragraph{Network architecture.}\label{sec:vqi2i-net}
The shared content encoder $E^c$, generators $\{G_X, G_Y\}$ and discriminators $\{D_X,D_Y\}$ in our model mostly follow the corresponding architectures proposed in VQGAN ~\cite{esser2021taming} but with two modifications: (1) We additionally concatenate four residual blocks with AdaIN layers (as what proposed in MUNIT~\cite{huang2018multimodal}) to the front of generators; (2) We replace the original normalization layers in discriminators with Instance normalization, as we train the whole model with a batch size of 1. For the style encoders $\{S_X, S_Y\}$, they are identical to the one used in MUNIT~\cite{huang2018multimodal}.

\paragraph{Adversarial loss.}
{\color{black} The adversarial loss is applied on the translated images $u$ and $v$ (cf. Eq.4) with respect to the real images $x\in X$ and $y\in Y$ respectively, where the discriminators (inherited from VQGAN) is used to matching their distributions ($u$ versus $x$; $v$ versus $y$). Specifically, we implement the adversarial loss as follow:
\begin{equation*}
\begin{split}
&L_{\text{adv}} = L_{D_X} + L_{D_Y},\\ 
&L_{D_X} = -[\log D_X(x) + \log (1-D_X(u))],\\
&L_{D_Y} = -[\log D_Y(y) + \log (1-D_Y(v))].
\end{split}
\end{equation*}
}

\subsection{Transformer}\label{sec:transformer-detail}
\paragraph{Setting of hyper-parameters and network architecture.}
For learning the autoregressive transformer model on the Yosemite dataset, we use the Adam optimizer with a batch size of 1, a learning rate of 0.00001, and exponential decay rates $(\beta_1, \beta_2) = (0.5, 0.999)$. 
The transformer we use is the same as the one used in VQGAN~\cite{esser2021taming}, which is identical to the GPT2 architecture~\cite{radford2019language}.
When in the testing phase, we set the parameters of sampling as follows: temperature $t=10$ and a top-$k$ cutoff at $k=2$.


\paragraph{Ordering difference between training and testing phases.}
In the training phase, we simply unfold the 2-dimensional quantized content representation of each training image on a row-major ordering (as the way VQGAN does) into a form of a discrete sequence and feed it into the transformer model for learning of content distribution. In other words, the transformer accesses the complete index sequence of an image at each time. 
In the testing phase, where we perform unconditional image generation and content extension, we design a square sliding window and feed only the indices in the current sliding window into the transformer (similarly on a row-major ordering as VQGAN). The transformer predicts a new content index only based on its previous indices within the sliding window, and the whole generation is done by moving the sliding window (in a left-to-right, top-to-bottom manner) and repeating this process. Here we use a sliding window with a size of $16\times16$.

\begin{figure*}[ht!]
\centering


    \begin{minipage}[b]{0.4\textwidth}
    \subfloat[Unconditional content generation followed by stylization with example-guided styles.]{%
		\includegraphics[width=\linewidth]{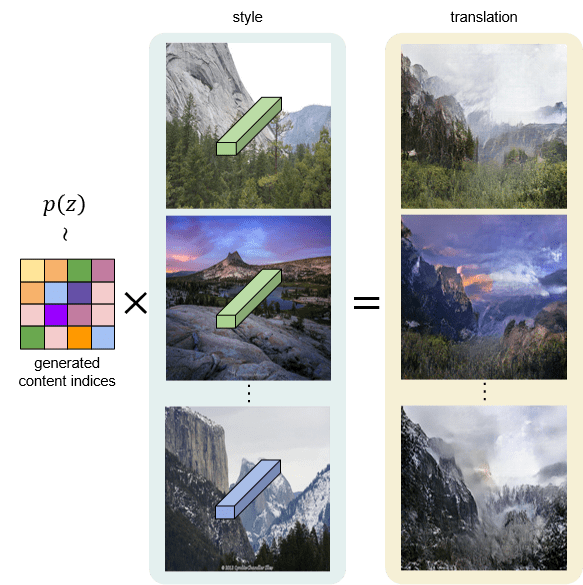}%
	}
    \end{minipage}
    \begin{minipage}[b]{0.55\textwidth}
    \subfloat[Diverse extension on an existing image.]{%
		\includegraphics[width=\linewidth]{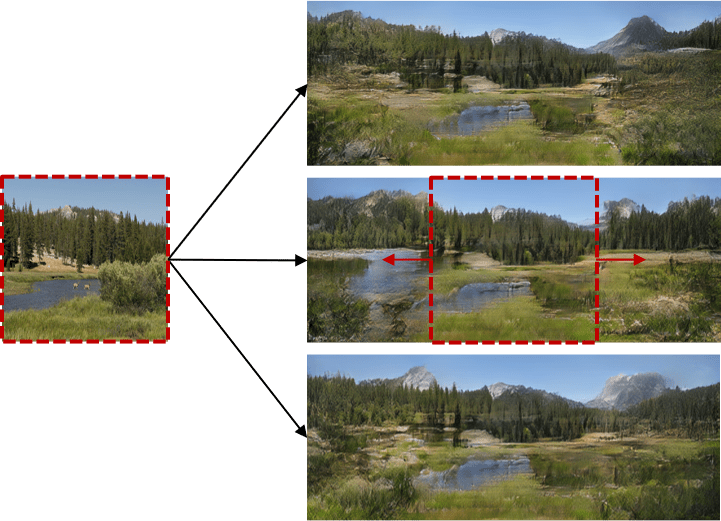}%
	}
    \end{minipage}
	\subfloat[Unconditional content generation with both translation/stylization and extension.]{%
		\includegraphics[width=0.95\linewidth]{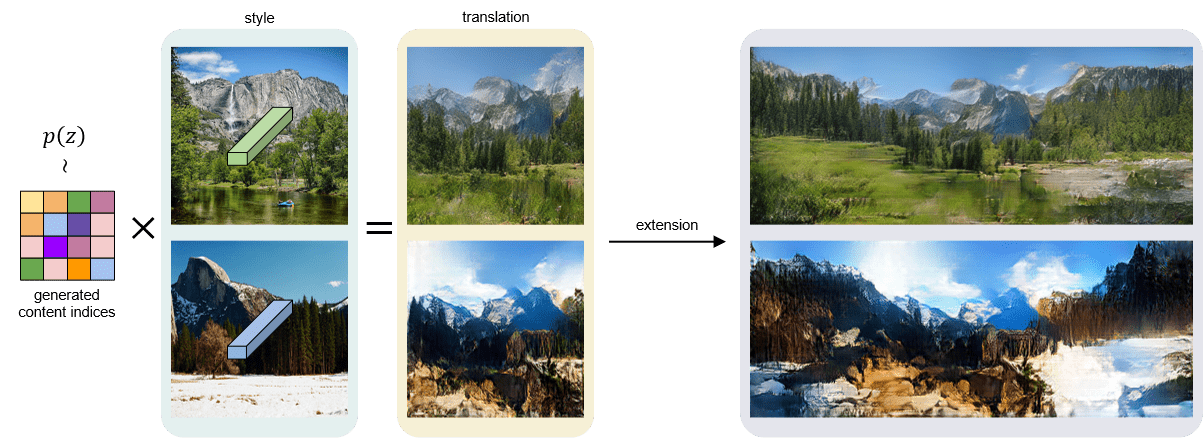}%
	}
    
    \caption{\textbf{Various applications with VQ-I2I.} (a) With generated content indices produced by the learned transformer model, we can combine some style features to synthesize diverse images. (b) Given an existing image, we are able to extend the diverse content on both sides. (c) We can combine unconditional content generation, diverse stylization, and extension together as a new application, which is easily achieved by our proposed method.}

\label{fig:application}
\end{figure*}

{\color{black}

\subsection{Applications}\label{sec:applications}
In addition to the transitional stylization generation described in the main paper, we further demonstrate several applications that our proposed method is able to unleash, as described in the following and shown in Figure~\ref{fig:application}:
\begin{itemize}
    \item \textbf{Unconditional content generation followed by stylization with example-guided styles.}  We first generate a sequence of content indices unconditionally from the learned transformer model and then modulate it with different styles from various domains.
    \item \textbf{Diverse extension on an existing image.} We utilize content indices from an existing image as the condition for the transformer model and generate the indices for the extended image region. We are able to produce diverse results of extension through having multiple samples drawn from the conditional distribution predicted by our transformer model. Figure~\ref{fig:application} (b) demonstrates the diversity on the results of such content extension.
    \item \textbf{Unconditional content generation with both translation/stylization and extension.} 
    We combine both content generation and extension with translation/stylization to showcase the flexibility on applications enabled by our proposed method as well as the variability of styles. We perform this combination in three steps: generate content indices unconditionally, modulate content indices with different style vectors, and apply extension.
    
\end{itemize}
}

\subsection{Unimodal VQ-I2I baseline (i.e., uni-VQ-I2I)}\label{sec:Uni-VQ-I2I}
\begin{figure*}[t]
    \centering
    \includegraphics[width=\linewidth]{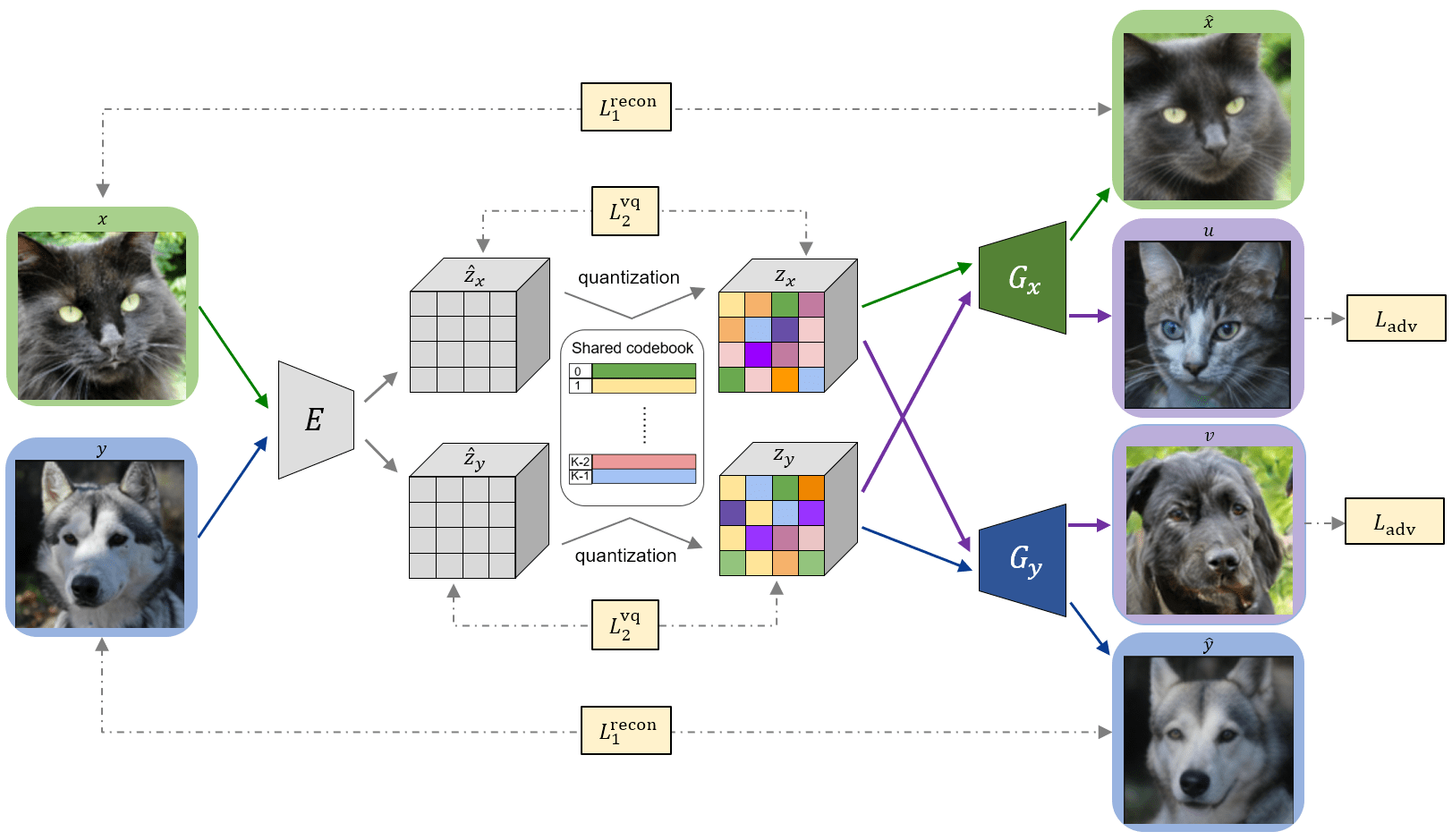}
    \caption{\textbf{Unimodal VQ-I2I.} The architecture illustration for the uni-VQ-I2I baseline.}
    \label{fig:Uni-VQ-I2I}
\end{figure*}

\indent As described in Section 4 of our main manuscript, we construct a uni-modal VQ-I2I variant as an additional baseline (denoted as uni-VQI2I), in which its latent space is not disentangled (i.e. there is no explicit separation between the content and style latent factors as our VQ-I2I). The architecture of uni-VQ-I2I is illustrated in Figure~\ref{fig:Uni-VQ-I2I}. Specifically, in such uni-VQ-I2I baseline, we assume that the domain-specific style information is implicitly modeled by the generators, thus the domain-specific style features are discarded.

As uni-VQ-I2I does not disentangle the latent space into the content and style parts, there are three main differences of unimodal VQ-I2I from multimodal VQ-I2I (i.e. our full model):
\begin{itemize}
    \item \textbf{Encoder.} uni-VQ-I2I only uses an public encoder $E$ to learn the joint latent space across domains.
    \item \textbf{Generators.} We remove the AdaIN normalization layers~\cite{huang2017arbitrary,huang2018multimodal} from $G_X, G_Y$, as there is only a single latent vector as input.
    \item \textbf{Loss function}. uni-VQ-I2I does not contain the style regression loss $L^{\mathrm{style}}_1$ and the content regression loss $L^{\mathrm{content}}_1$, and we modify the full loss function to
    \begin{equation}
    \begin{split}
    L_{D} &= L_\mathrm{adv},\\
    L_{E, Z, G} &= -\lambda_\mathrm{adv} L_\mathrm{adv} + \lambda^\mathrm{recon}_1 L^\mathrm{recon}_1 + \lambda_\mathrm{vq} L_\mathrm{vq}.
    \end{split}
    \label{eq:univq-loss}
    \end{equation}
    
\end{itemize}
\paragraph{Network architecture}
Similar to what has been described in Section~\ref{sec:vqi2i-net}, the encoder, generators and discriminators in the architecture of uni-VQ-I2I are mostly inherited from VQGAN~\cite{esser2021taming}. 

\begin{figure}[t]
    \centering
    \includegraphics[width=.95\linewidth]{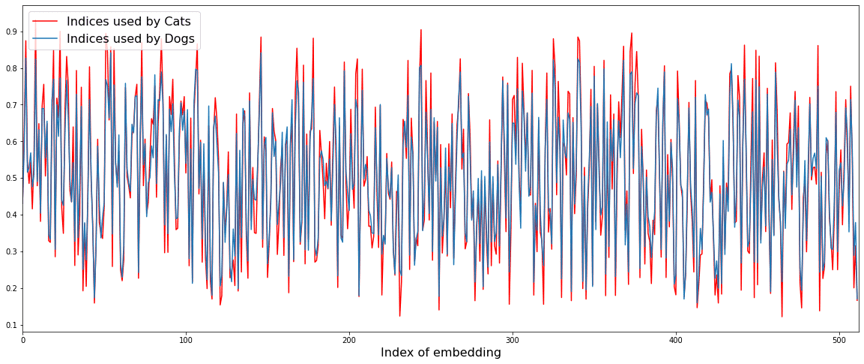}
    \caption{\textbf{Embeddings/codes used by training uni-VQ-I2I model for the translation between Cats and Dogs in AFHQ dataset.} It is observable that the embeddings/codes used by these two domains are highly overlapped. Noting the numerical range on $y$-axis has been normalized with respect to total number of training images of each domain.}
    \label{fig:indices}
\end{figure}

\paragraph{Uni-VQ-I2I exploration.}
To verify whether the generators in uni-VQ-I2I are able to handle the translation without having the disentangled representations, we record the total number of embeddings/codes being used in two domains.
Take the training set of AFHQ dataset as an example, Figure~\ref{fig:indices} reveals that the used embeddings/codes of cat and dog images are highly overlapped.
This fact indicates that the model tends to learn a general latent representation for the codebook, rather than using specific codes for specific domains.
That is, the translation is implicitly handled by domain-specific generators.
Besides, all the codes in the codebook (number of embeddings = $512$) are used.

We provide additional qualitative results in Figure~\ref{fig:afhq-res} and~\ref{fig:yosemite-res} (from AFHQ and Yosemite datasets respectively) to make comparison between VQ-I2I and uni-VQ-I2I, where we can observe that:   
although uni-VQ-I2I seems to be capable of addressing the I2I task, the content of the translation results is not consistent with the corresponding source images.

\subsection{Details of compared baselines.}
For the I2I baselines, we follow their official code and default settings on training and hyper-parameters (noting that for U-GAT-IT~\cite{kim2019u} we adopt its \textit{light} version out of consideration for computational cost).
As DRIT~\cite{lee2018diverse}, MUNIT~\cite{huang2018multimodal} and BicycleGAN~\cite{zhu2017toward} are multi-modal models, we sample one translation result for every input image for FID computation.\par
{\color{black}
Regarding the baseline of image extension, Boundless~\cite{teterwak2019boundless}, we adopt the implementation from \url{https://github.com/recong/Boundless-in-Pytorch}, in which we train the models without modifying any hyperparameters to perform horizontal extension for 50\% and 75\% toward the right-hand side as mentioned in the main paper. 
%
}
\section{Additional Experiments}

\subsection{\textbf{Multimodal Translation on Cat$\rightarrow$Wild.}}
\begin{figure}[h]
    \centering
    \setlength\tabcolsep{1.5pt} 
    \begin{tabular}{c:c}
    Input & Multimodal translation \\
    \includegraphics[height=.5\linewidth]{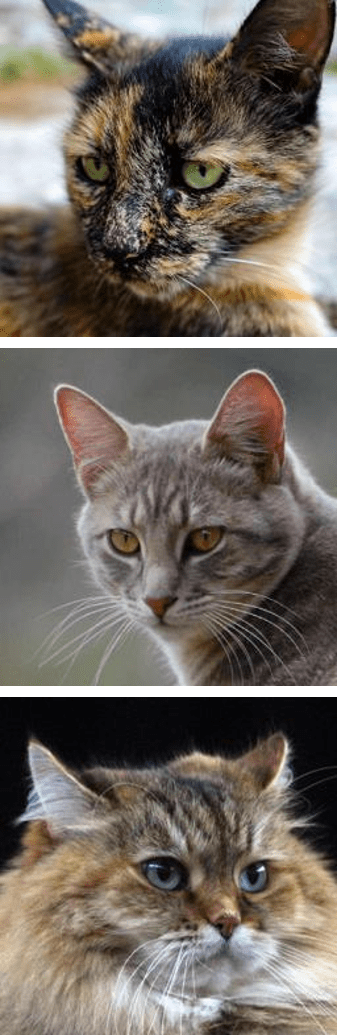} &
    \includegraphics[height=.5\linewidth]{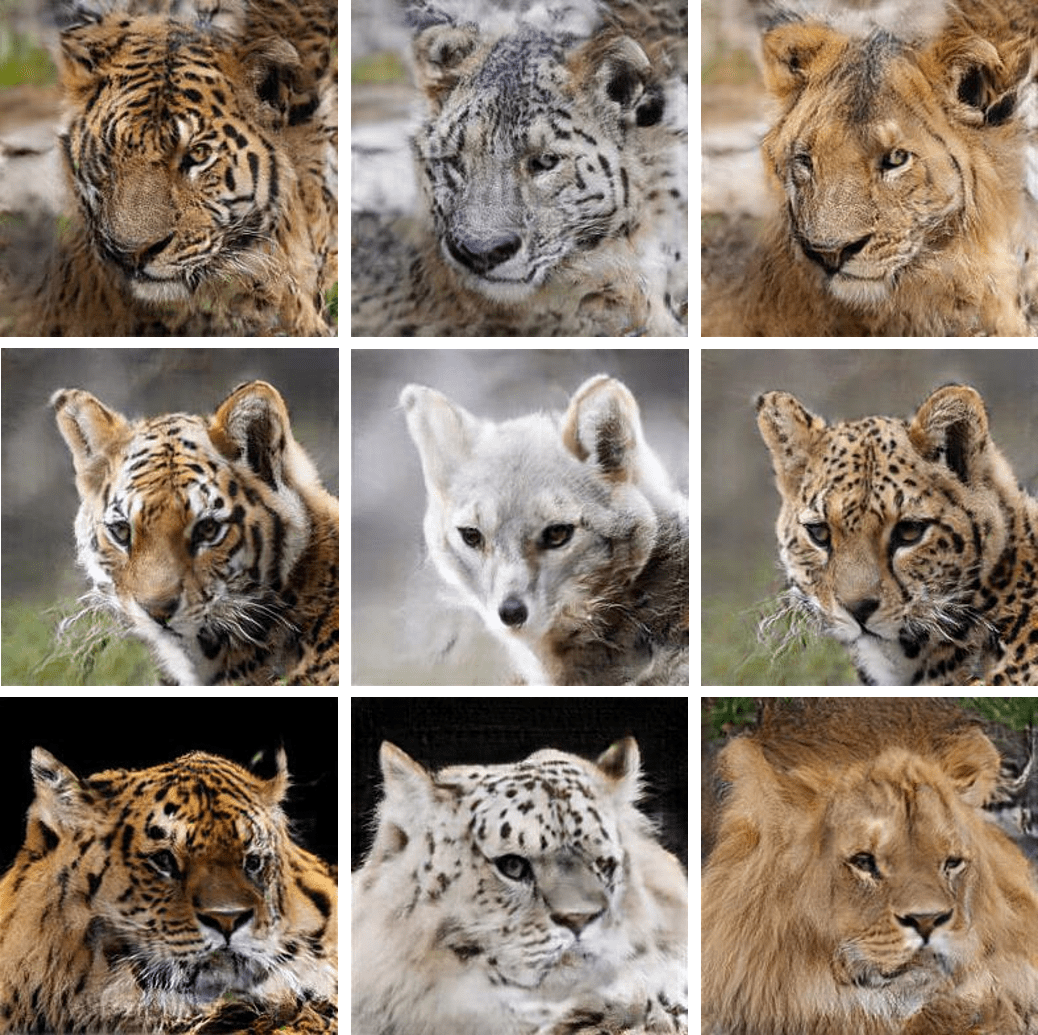} \\
    \end{tabular}
\caption{\textbf{Multimodal translation on cat$\rightarrow$wild.} We present the inter-domain translations on cat$\rightarrow$wild. Our VQ-I2I model is able to generate different categories of wide animals in the target domain.
}
\label{fig:wild}
\end{figure}

In our main manuscript, we have presented the intra- and inter-domain multimodal translation.
Here we also provide more experimental results on the AFHQ dataset for translating from cats to wildlife animals.
The wildlife animals in AFHQ dataset contains various animals, such as lions, tigers and leopards.
We present the inter-domain multimodal results in Figure~\ref{fig:wild}.
The generator is able to learn a general representation of various styles in the same domain.

\begin{figure*}[t]
    \centering
   \includegraphics[width=.95\linewidth]{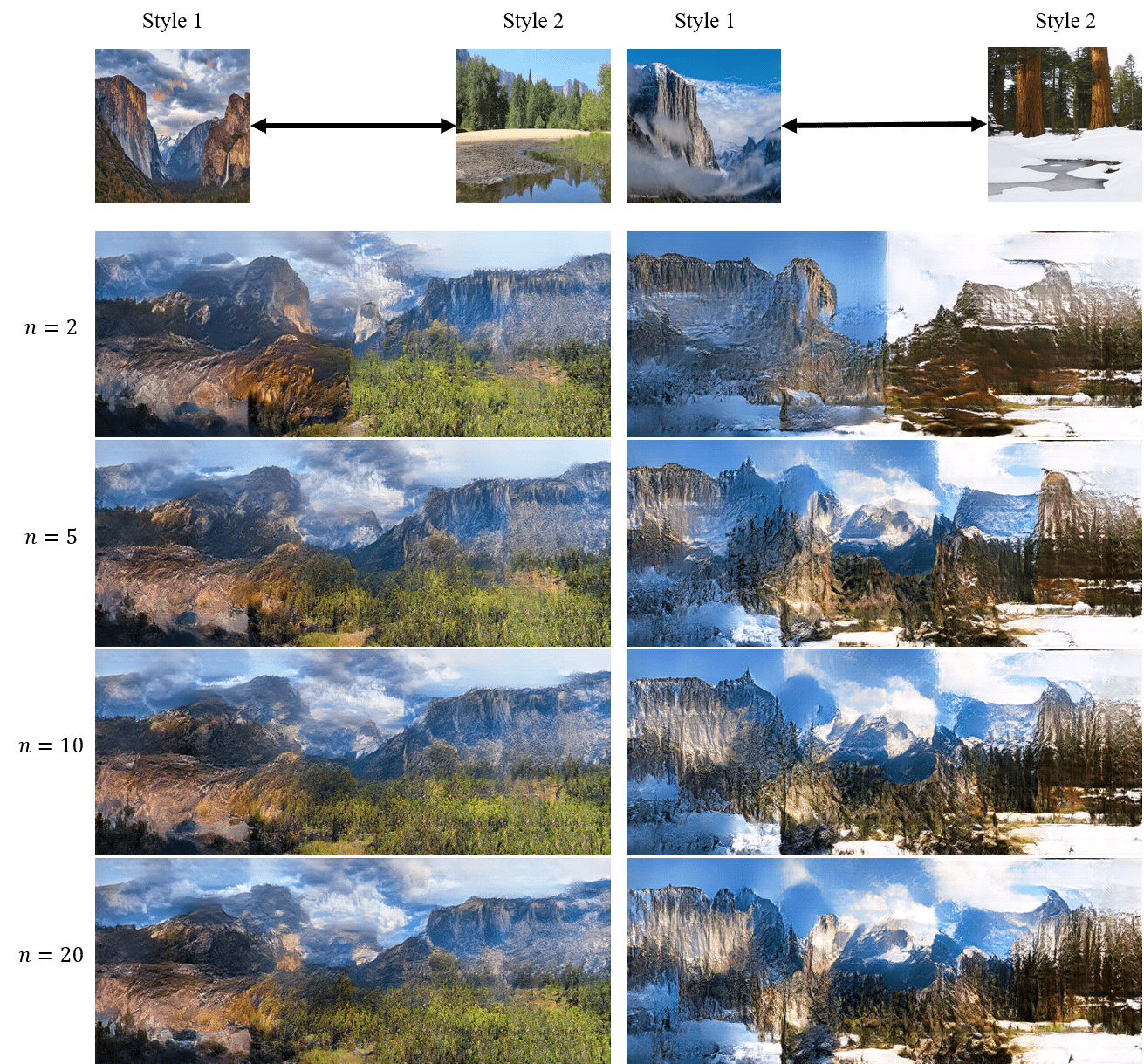}
    \caption{{\color{black}\textbf{Experiments to have different number of splits on the content map to perform transitional stylization.} The first row shows the two referenced styles, and the second to the last rows are the transitional results when using 2, 5, 10 and 20 splits, respectively.}} 
    \label{fig:diff-splits}
\end{figure*}

{\color{black}
\subsection{Having Different Number of Splits for Transitional Stylization}
In Figure~\ref{fig:diff-splits} we present the results of transitional stylization with using different numbers of splits (noting that we partition the content map horizontally into several equal splits and modulate different parts of the content map independently with different proportions by mixing the two styles, as described in Sec. 4.3 of the main manuscript). We observe that when the number of splits increase, the transitional stylization gets more smooth to gradually change from one style to another.  
}

{\color{black}
\subsection{Baseline via sequential combination of SOTA methods.}
We build a baseline based on a sequential combination of StyleGAN2 (generation), U-GAT-IT (translation), and Boundless (extension) to make a comparison between VQ-I2I and a sequential combination of SOTA mothods. With conducting an experiment of firstly generating 100 landscape images of size 256$\times$256, translating them to summer styles, and finally extending the width for 128 pixels toward the right-hand side, our VQ-I2I is able to provide superior performance with FID 107.62 than such baseline 
with FID 128.73 (w.r.t. summer images from Yosemite dataset).}

\subsection{More comparisons with recent papers on unconditional generation and extension tasks.}
{\color{black} We include more recent baselines on unconditional generation (i.e. StyleGAN2~\cite{karras2020analyzing}) and extension (i.e. InfinityGAN~\cite{lin2021infinitygan}). For unconditional generation, StyleGAN2, our VQ-I2I, and VQGAN achieve FID 106.35, 127.31, and 127.84 respectively;
For image extension, following InfinityGAN's setting (i.e. given images of size 256$\times$128 and extending them to 256$\times$256), InfinityGAN, our VQ-I2I, and Boundless achieve FID 143.97, 109.86, and 101.68 respectively (Noting FID scroes above are all evaluated on 100 generated/extended images w.r.t. Yosemite dataset). Though the main focus of our VQ-I2I is to handle multiple tasks in a unified framework instead of targeting the state-of-the-art performance, it still provides comparable performance with the recent works on generation or extension.}

\section{Additional Results}

\subsection{Image-to-Image Translation}
\begin{figure*}[t]
    \centering
    
    \setlength\tabcolsep{1pt} 
    \begin{tabular}{c:ccccccc}
    \tiny Input & \tiny Ours & \tiny uni-VQ-I2I & \tiny CycleGAN & \tiny DRIT & \tiny MUNIT & \tiny U-GAT-IT & \tiny CUT \\
    \includegraphics[width=.12\linewidth]{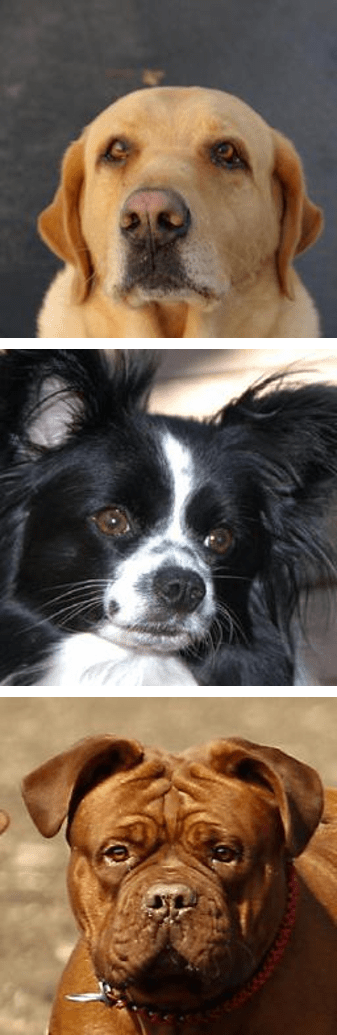} &
    \includegraphics[width=.12\linewidth]{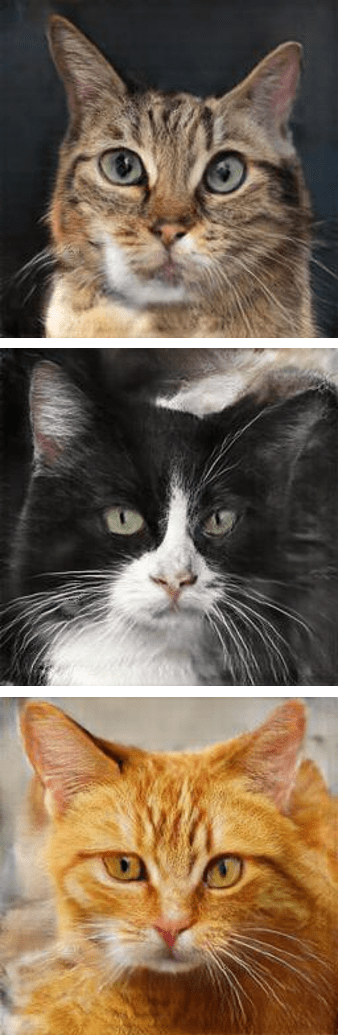} &
    \includegraphics[width=.12\linewidth]{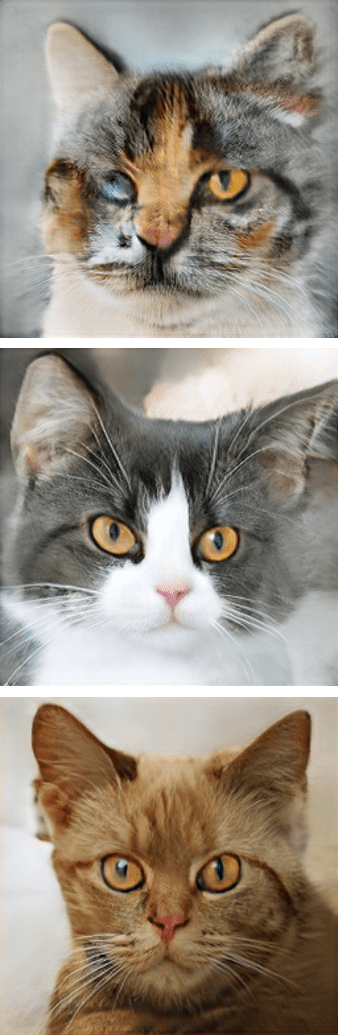} &
    \includegraphics[width=.12\linewidth]{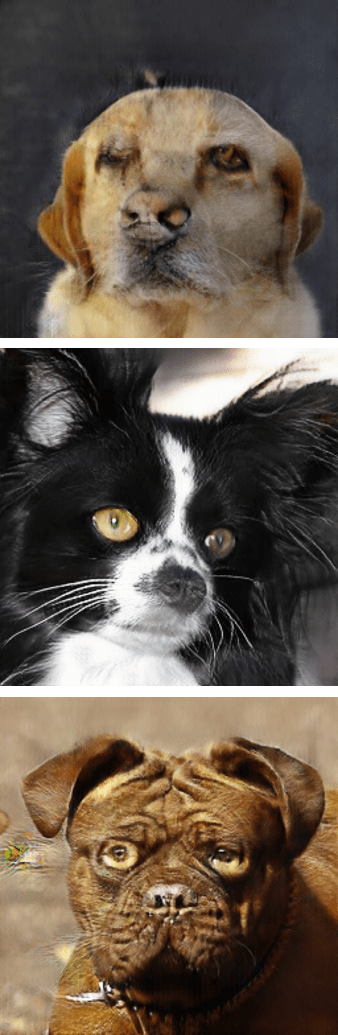} &
    \includegraphics[width=.12\linewidth]{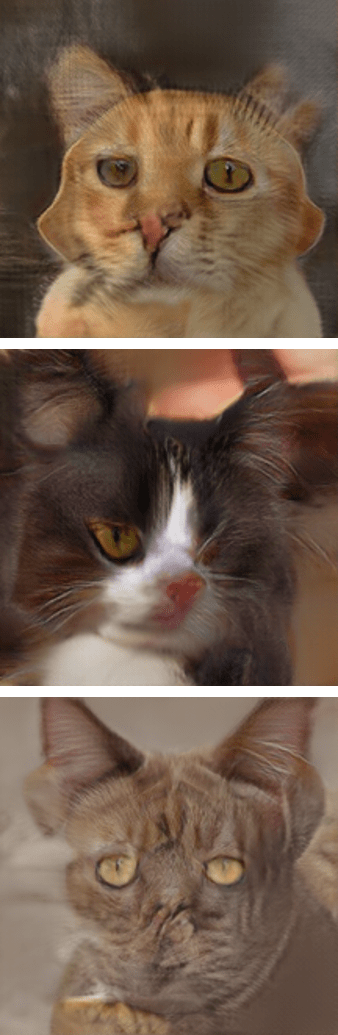} &
    \includegraphics[width=.12\linewidth]{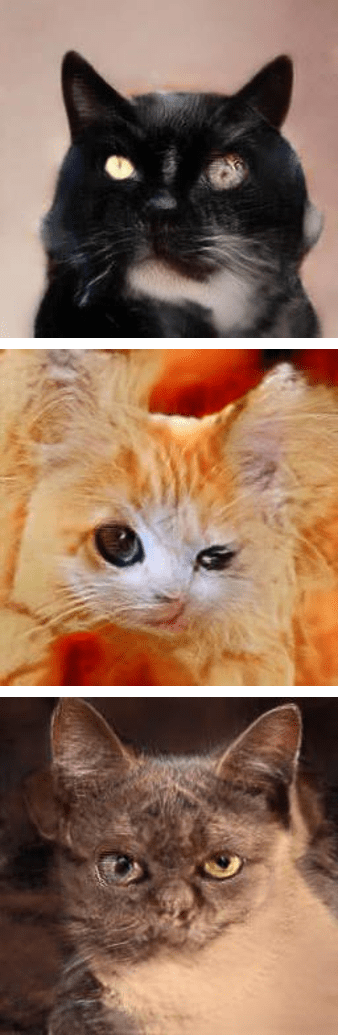} &
    \includegraphics[width=.12\linewidth]{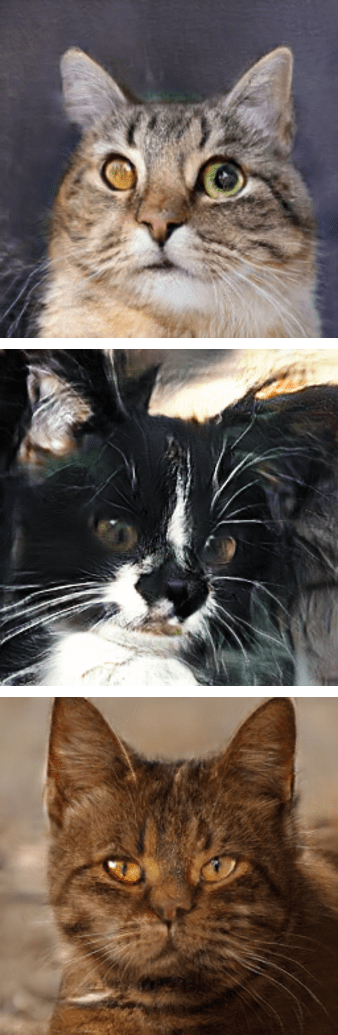} &
    \includegraphics[width=.12\linewidth]{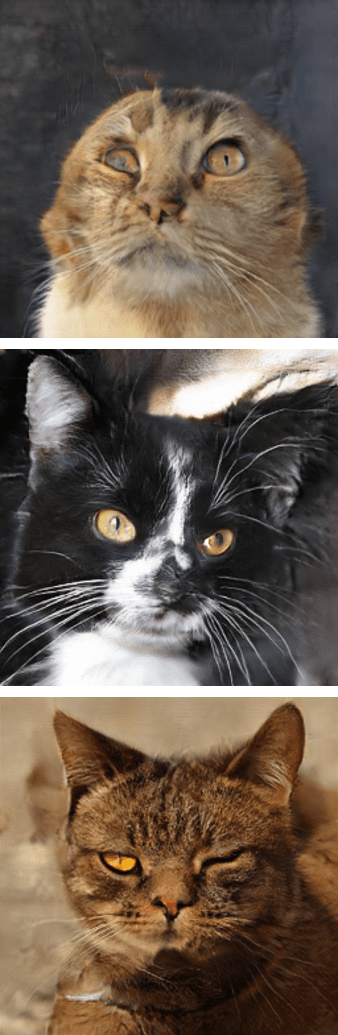} \\
    \midrule
    \includegraphics[width=.12\linewidth]{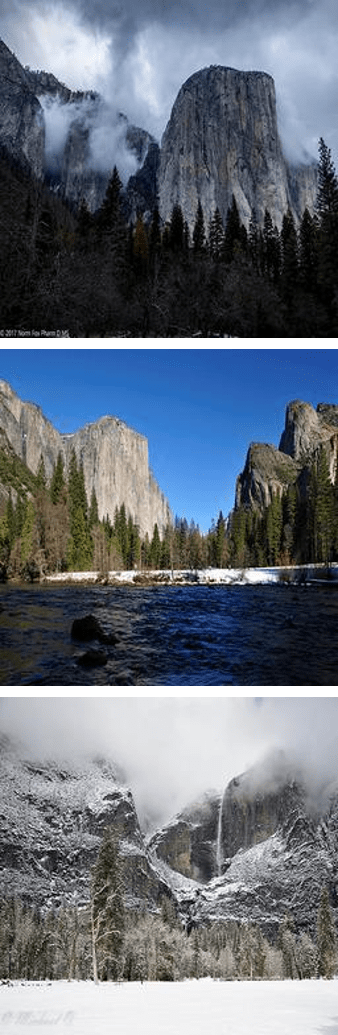} &
    \includegraphics[width=.12\linewidth]{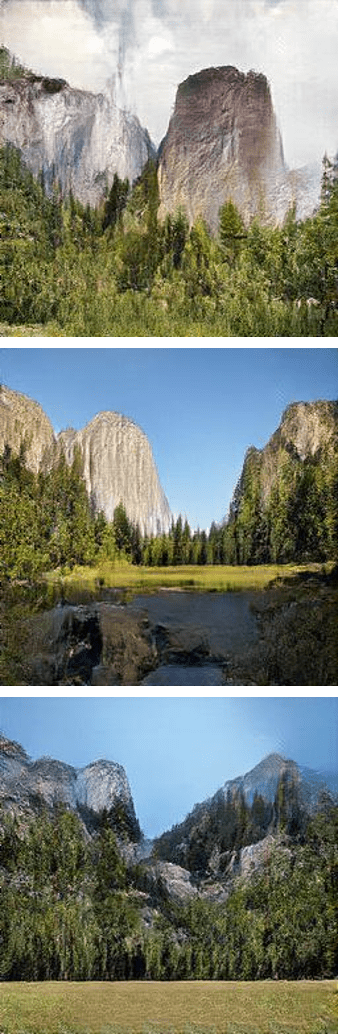} &
    \includegraphics[width=.12\linewidth]{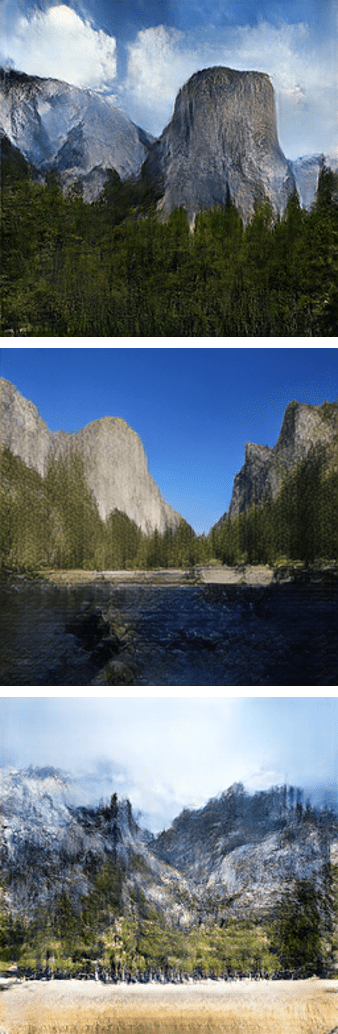} &
    \includegraphics[width=.12\linewidth]{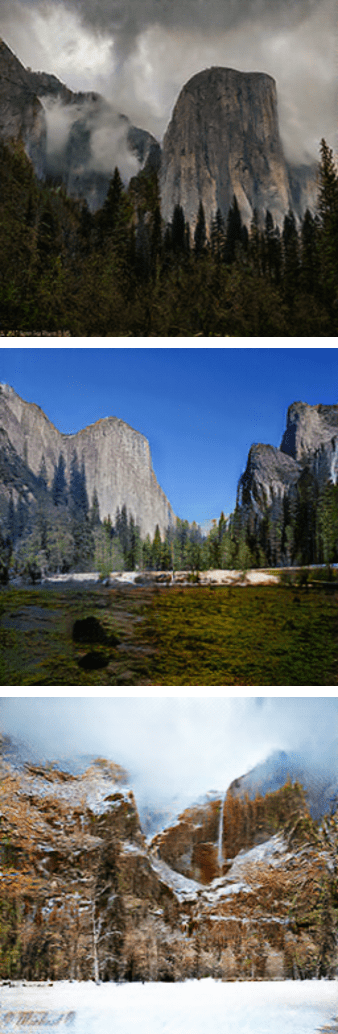} &
    \includegraphics[width=.12\linewidth]{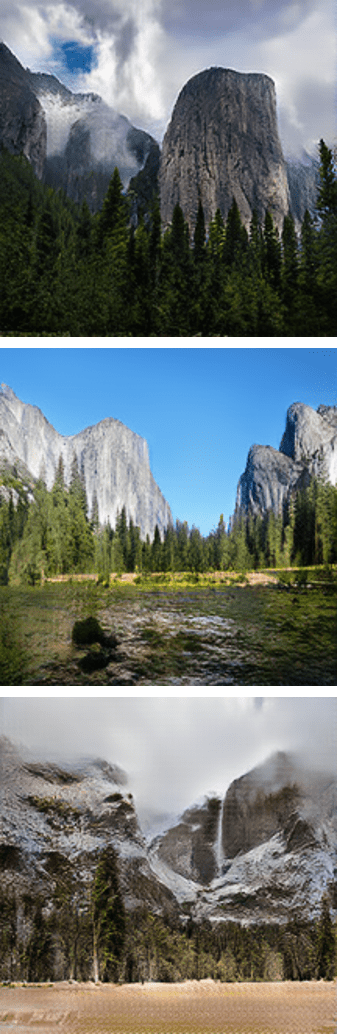} &
    \includegraphics[width=.12\linewidth]{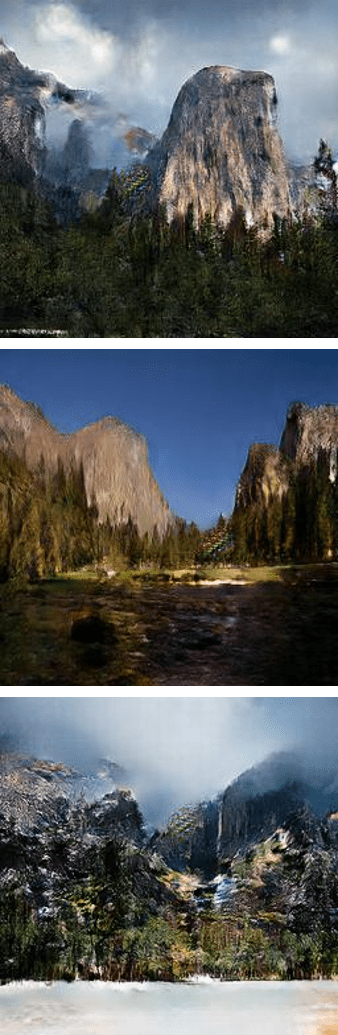} &
    \includegraphics[width=.12\linewidth]{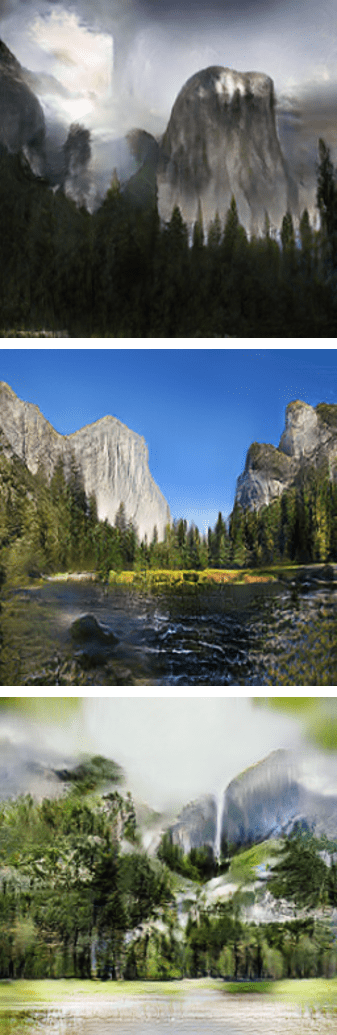} &
    \includegraphics[width=.12\linewidth]{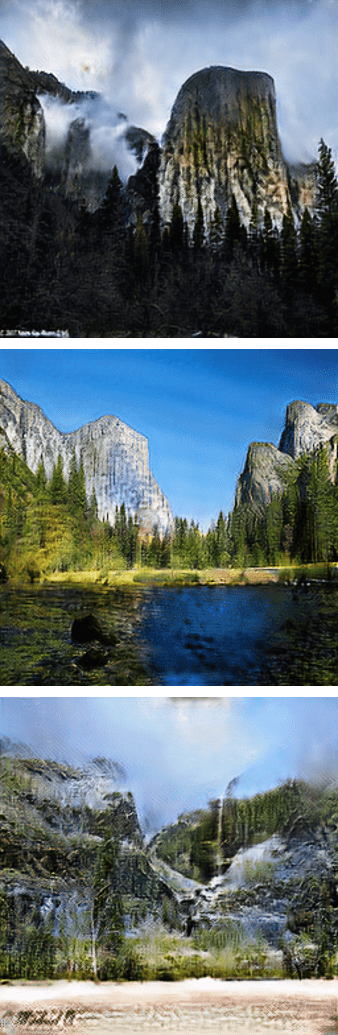} \\
    \midrule
    \includegraphics[width=.12\linewidth]{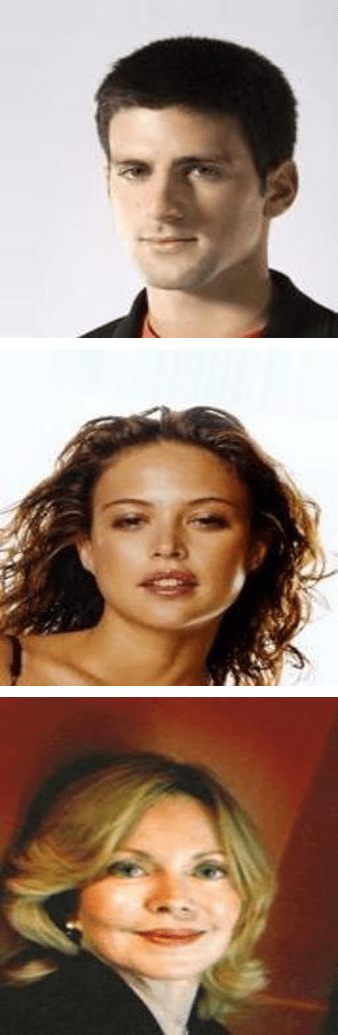} &
    \includegraphics[width=.12\linewidth]{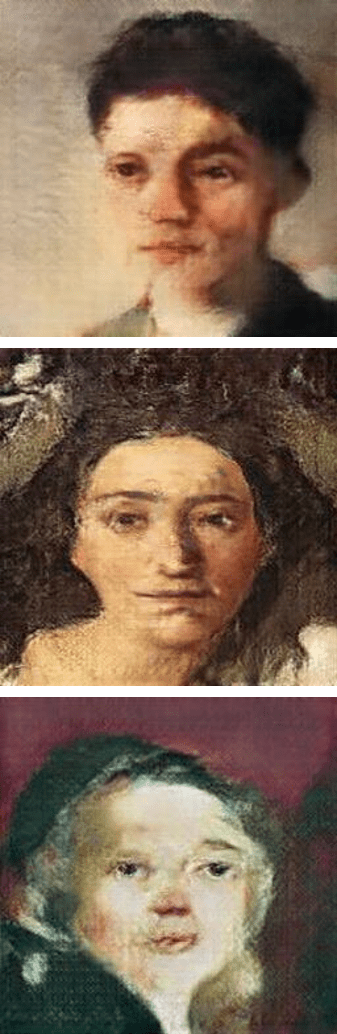} &
    \includegraphics[width=.12\linewidth]{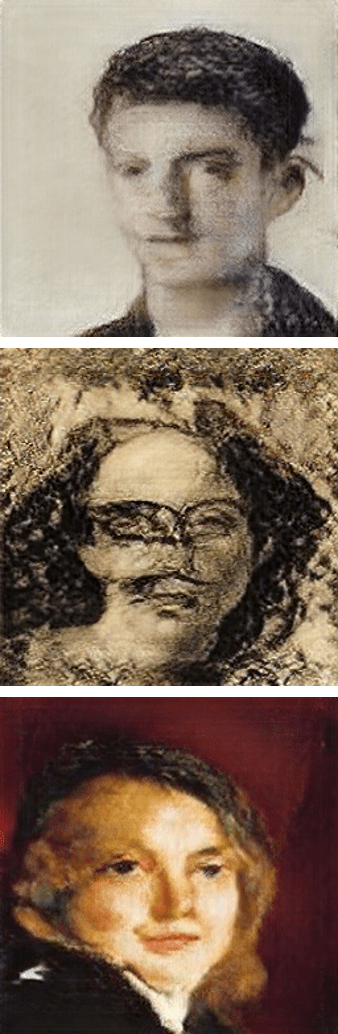} &
    \includegraphics[width=.12\linewidth]{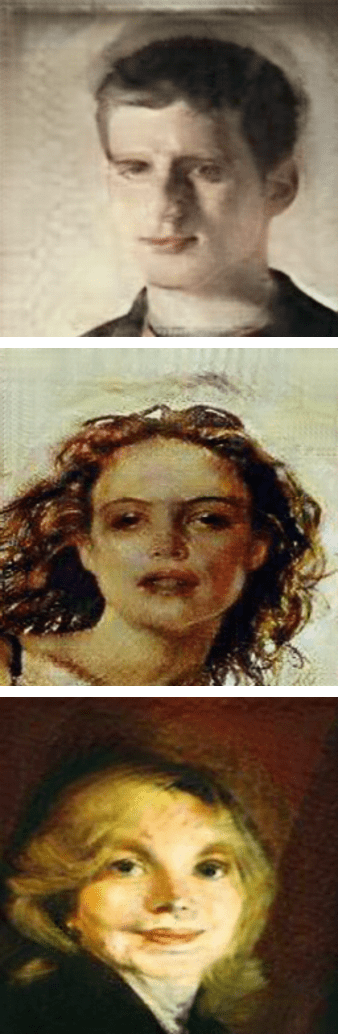} &
    \includegraphics[width=.12\linewidth]{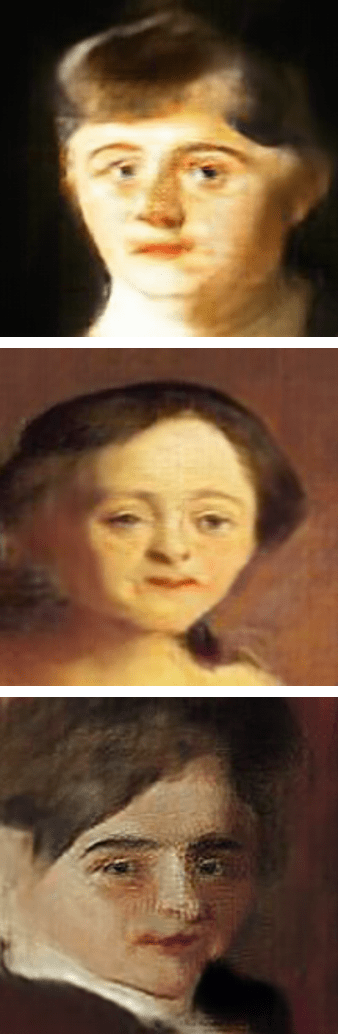} &
    \includegraphics[width=.12\linewidth]{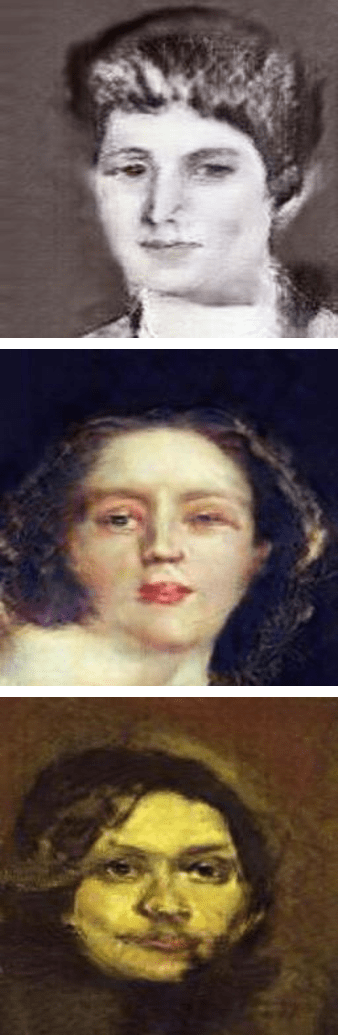} &
    \includegraphics[width=.12\linewidth]{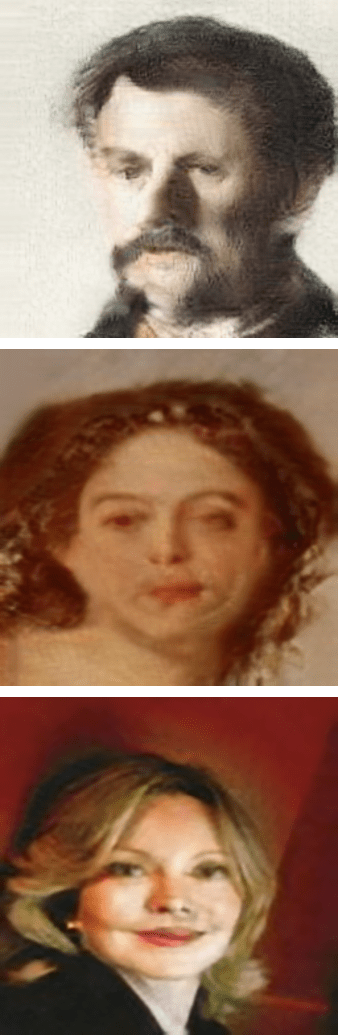} &
    \includegraphics[width=.12\linewidth]{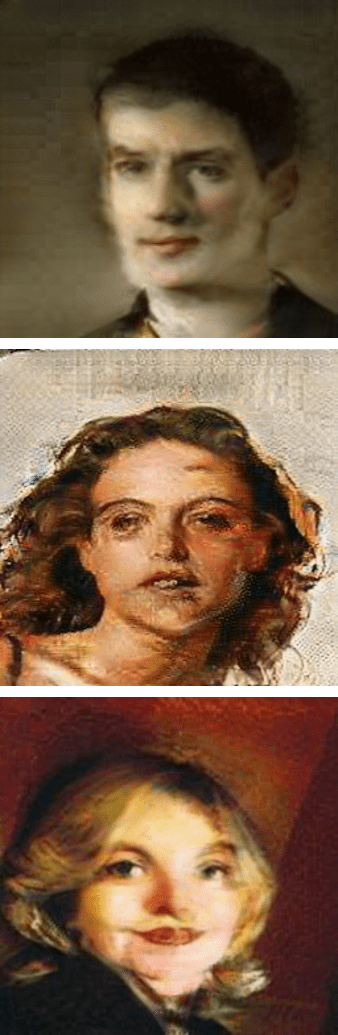} \\
    
    \end{tabular}
\caption{\textbf{More qualitative comparisons with conventional image-to-image translation methods.}
We provide qualitative examples of the translation results produced by various methods, trained on unpaired datasets. The left-most column shows the input images in the source domain. The other seven columns show the corresponding translated images in the target domain. Every three rows from top to bottom are: dog$\rightarrow$cat, winter$\rightarrow$summer, and photo$\rightarrow$portrait.
}
\label{fig:baseline}
\end{figure*}

\begin{figure*}[t]
    \centering
    \setlength\tabcolsep{1.5pt} 
    \subfloat[]{
        \begin{tabular}{c:ccc}
        Input & Ours & uni-VQ-I2I \\
        \includegraphics[width=.15\linewidth]{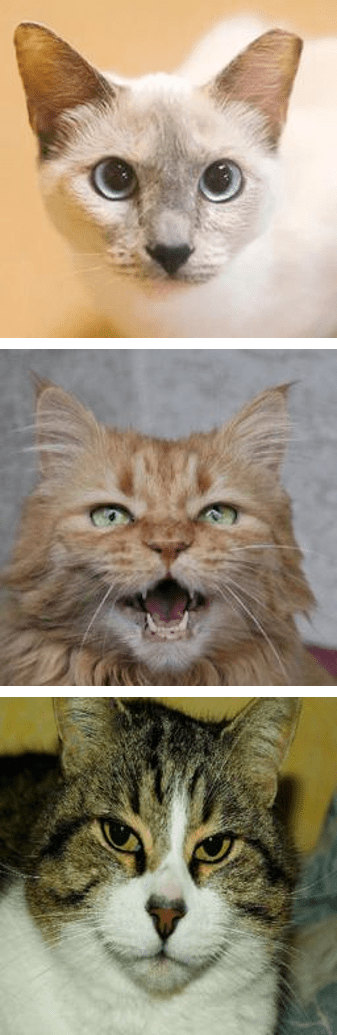} &
        \includegraphics[width=.15\linewidth]{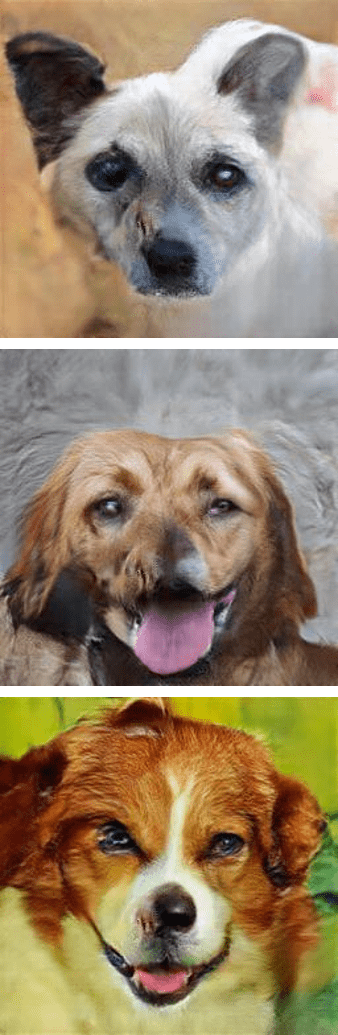} &
        \includegraphics[width=.15\linewidth]{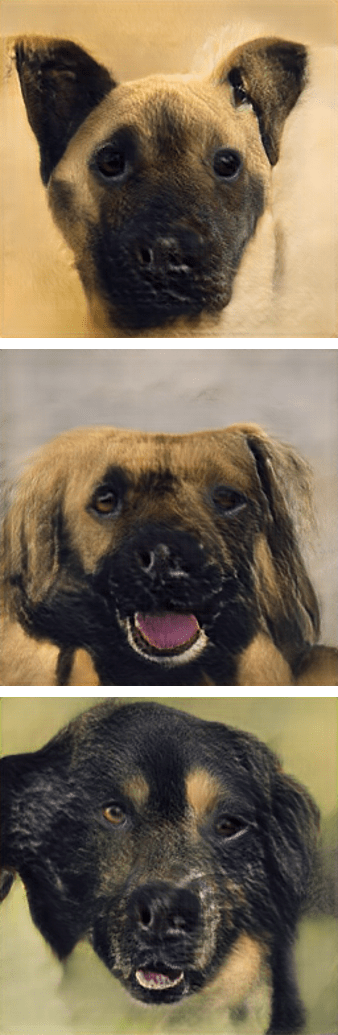} \\
        \end{tabular}
    }
    \subfloat[]{
        \begin{tabular}{c:ccc}
        Input & Ours & uni-VQ-I2I \\
        \includegraphics[width=.15\linewidth]{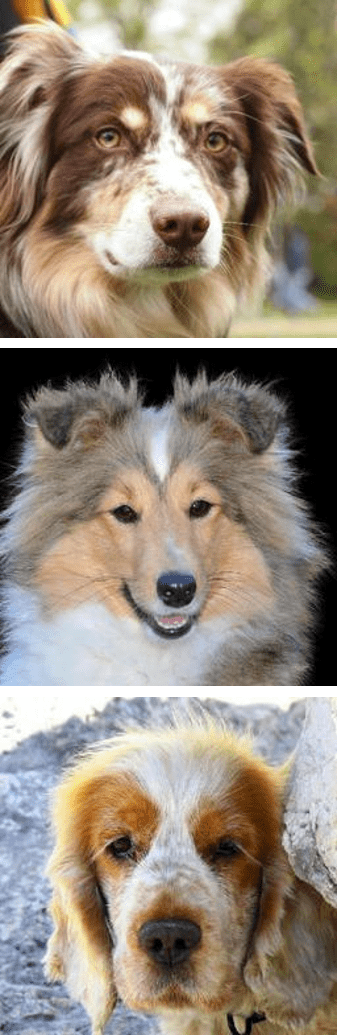} &
        \includegraphics[width=.15\linewidth]{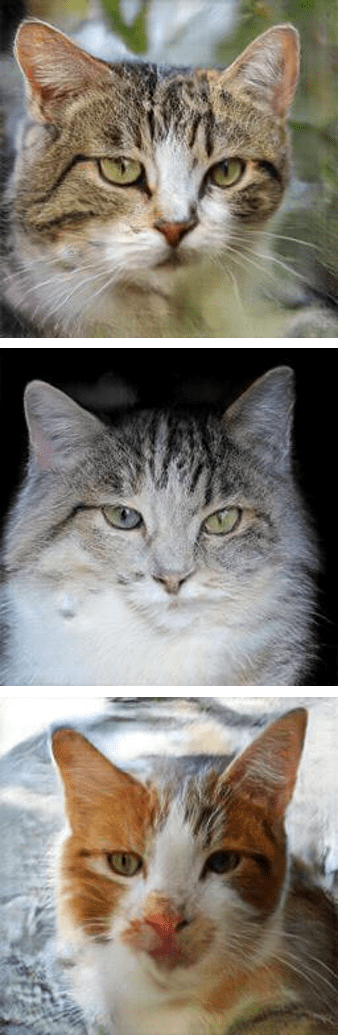} &
        \includegraphics[width=.15\linewidth]{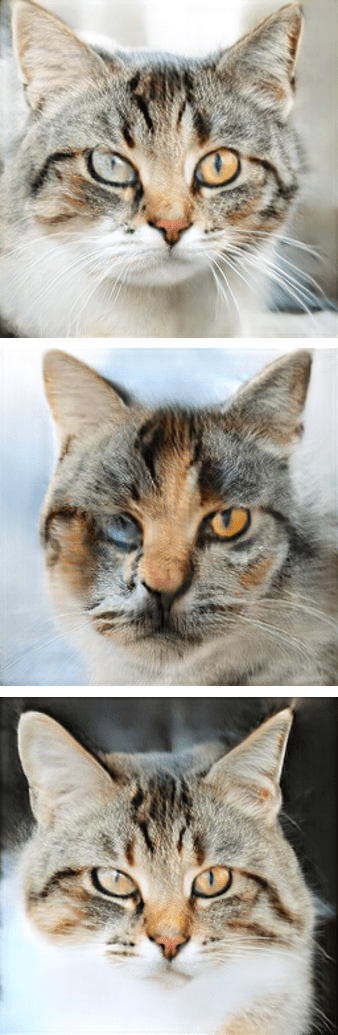} \\
        \end{tabular}
    }
\caption{\textbf{Qualitative comparison between our VQ-I2I and uni-VQ-I2I baseline on AFHQ dataset.}
Three columns on the left show the translation cat$\rightarrow$dog, while three columns on the right show the translation dog$\rightarrow$cat.
}
\label{fig:afhq-res}
\end{figure*}

\begin{figure*}[t]
    \centering
    \setlength\tabcolsep{1.5pt} 
    
    \subfloat[]{
        \begin{tabular}{c:ccc}
        input & Ours & uni-VQ-I2I \\
        \includegraphics[width=.15\linewidth]{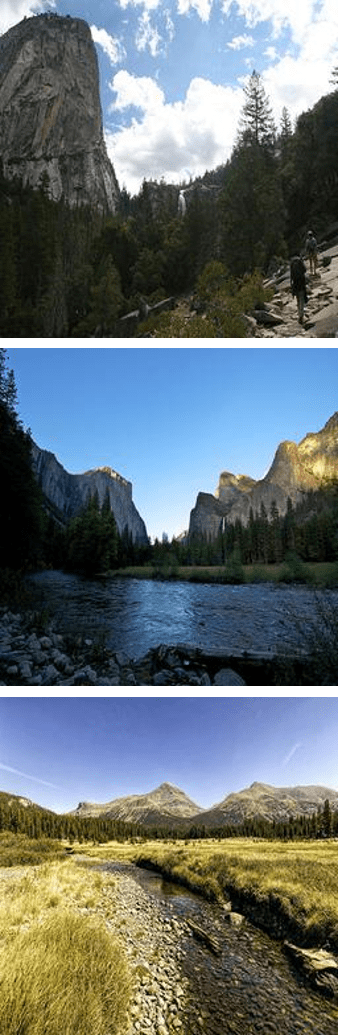} &
        \includegraphics[width=.15\linewidth]{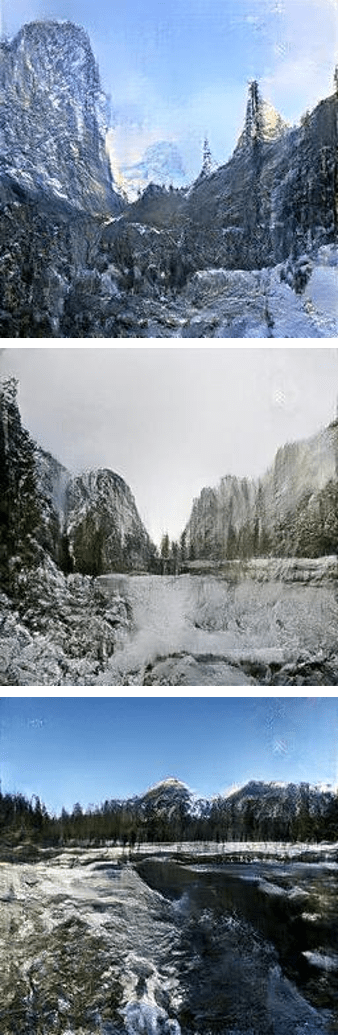} &
        \includegraphics[width=.15\linewidth]{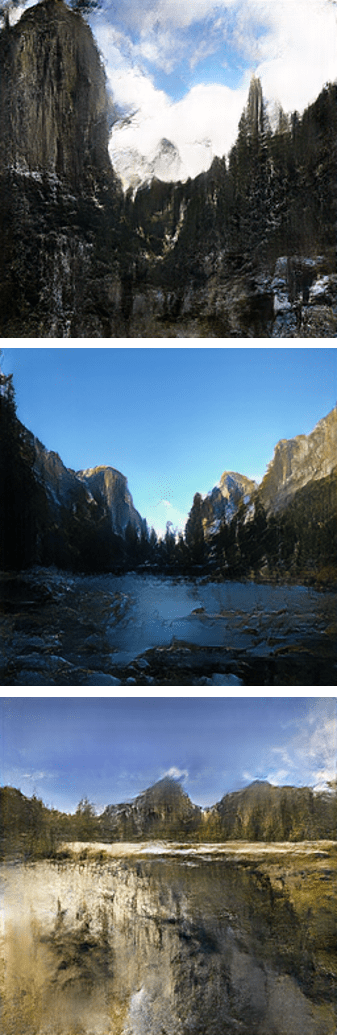} \\
        \end{tabular}
    }
    \subfloat[]{
        \begin{tabular}{c:ccc}
        input & Ours & uni-VQ-I2I \\
        \includegraphics[width=.15\linewidth]{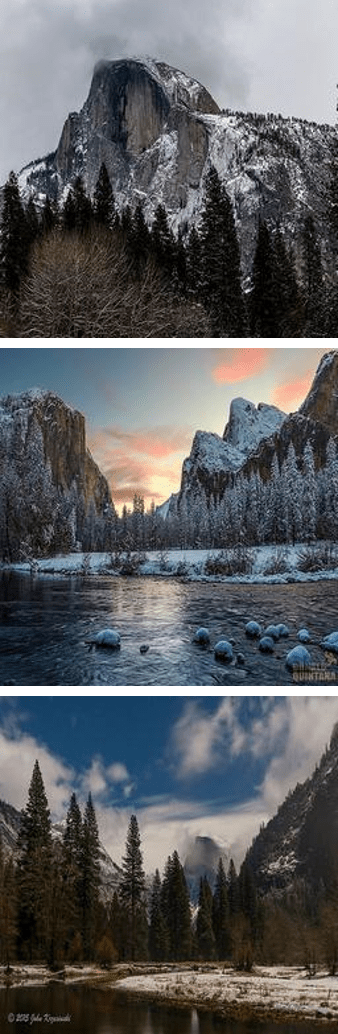} &
        \includegraphics[width=.15\linewidth]{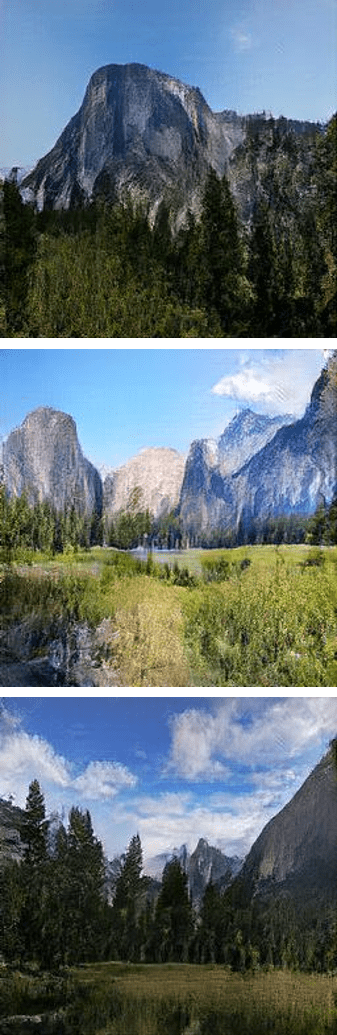} &
        \includegraphics[width=.15\linewidth]{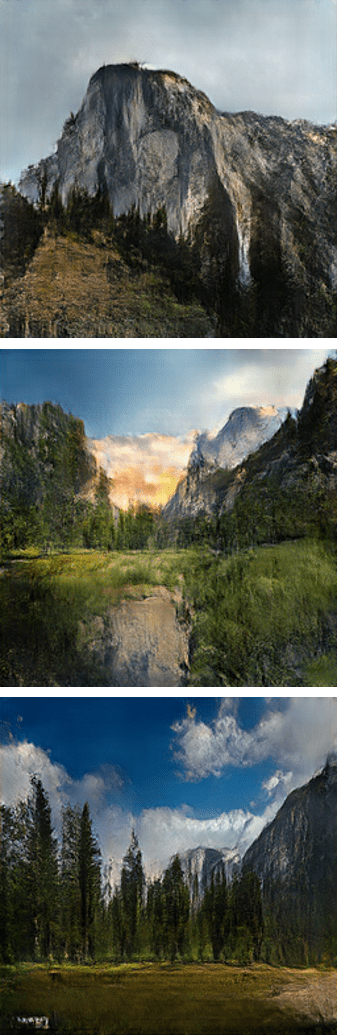} \\
        \end{tabular}
    }
\caption{\textbf{Qualitative comparison between our VQ-I2I and uni-VQ-I2I baseline on Yosemite dataset.}
 Three columns on the left show the translation summer$\rightarrow$winter, while three columns on the right show the translation winter$\rightarrow$summer.
}
\label{fig:yosemite-res}
\end{figure*}

In Figure~\ref{fig:baseline}, we present additional results for dog$\rightarrow$cat, winter$\rightarrow$summer, and photo$\rightarrow$portrait, obtained by various methods. Besides, we show the translation results of VQ-I2I and uni-VQ-I2I for both directions on shape-variant (AFHQ) and shape-invariant (Yosemite) datasets in Figure~\ref{fig:afhq-res}
and~\ref{fig:yosemite-res}, respectively.


\subsection{Applications}
\begin{figure*}[t]
    \centering
    \setlength\tabcolsep{1.5pt} 
    \begin{tabular}{c:cc}
    Uncond. generation & translation & translation + extension \\
    \includegraphics[height=.40\linewidth]{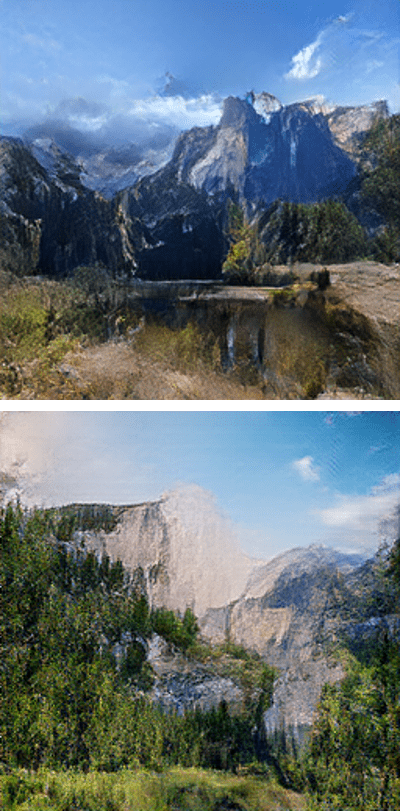} &
    \includegraphics[height=.40\linewidth]{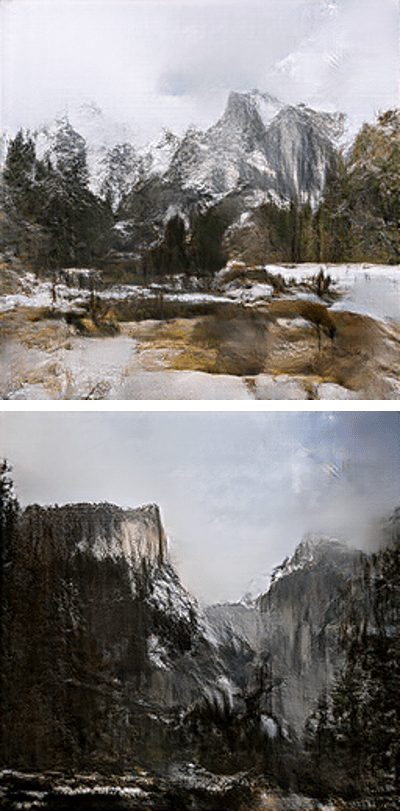} & 
    \includegraphics[height=.40\linewidth]{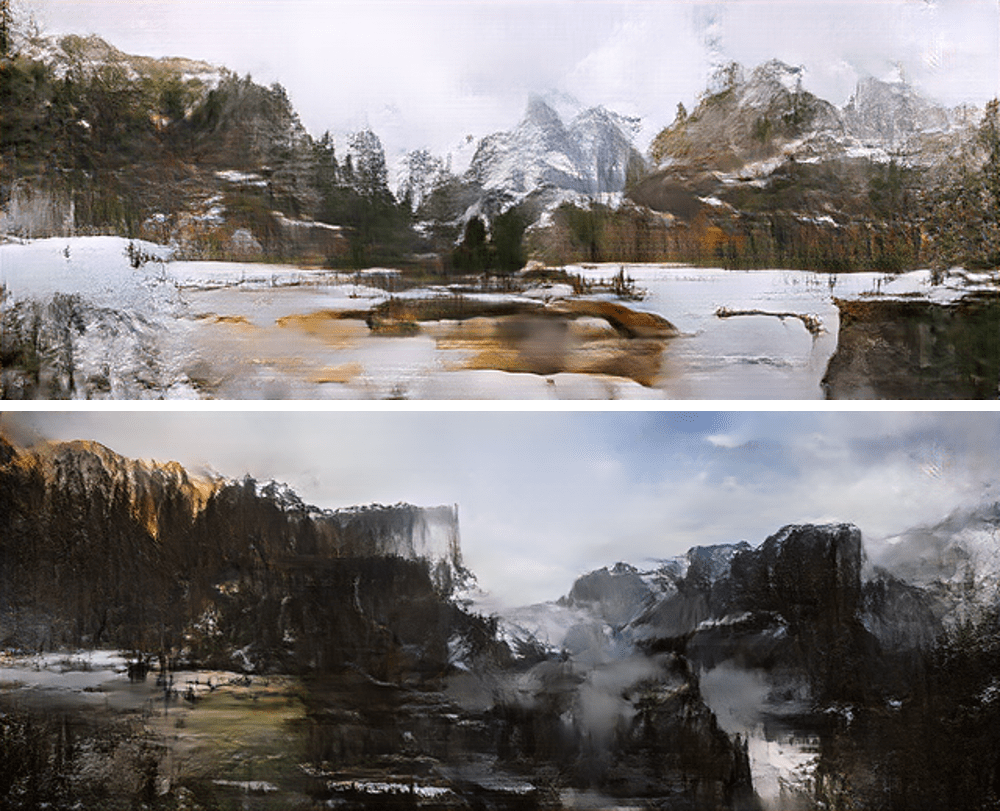} \\
    \midrule
    \includegraphics[height=.40\linewidth]{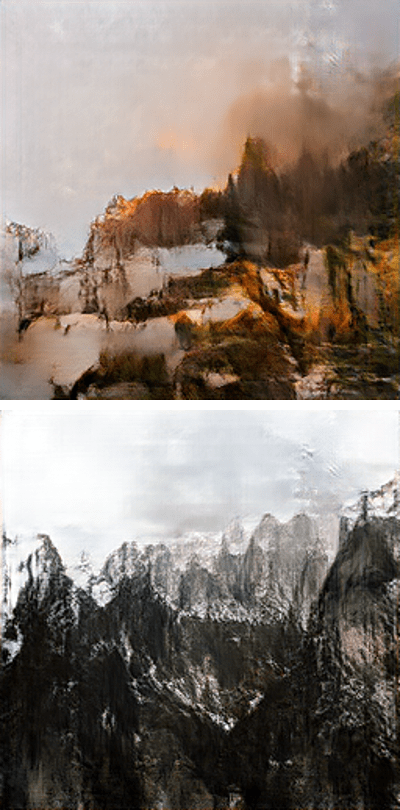} &
    \includegraphics[height=.40\linewidth]{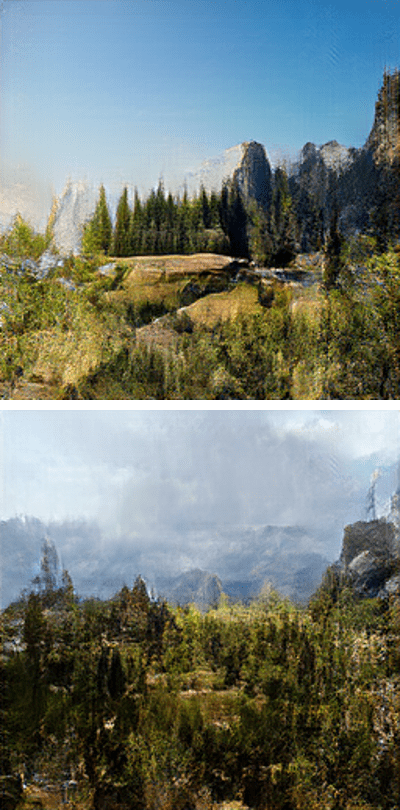} & 
    \includegraphics[height=.40\linewidth]{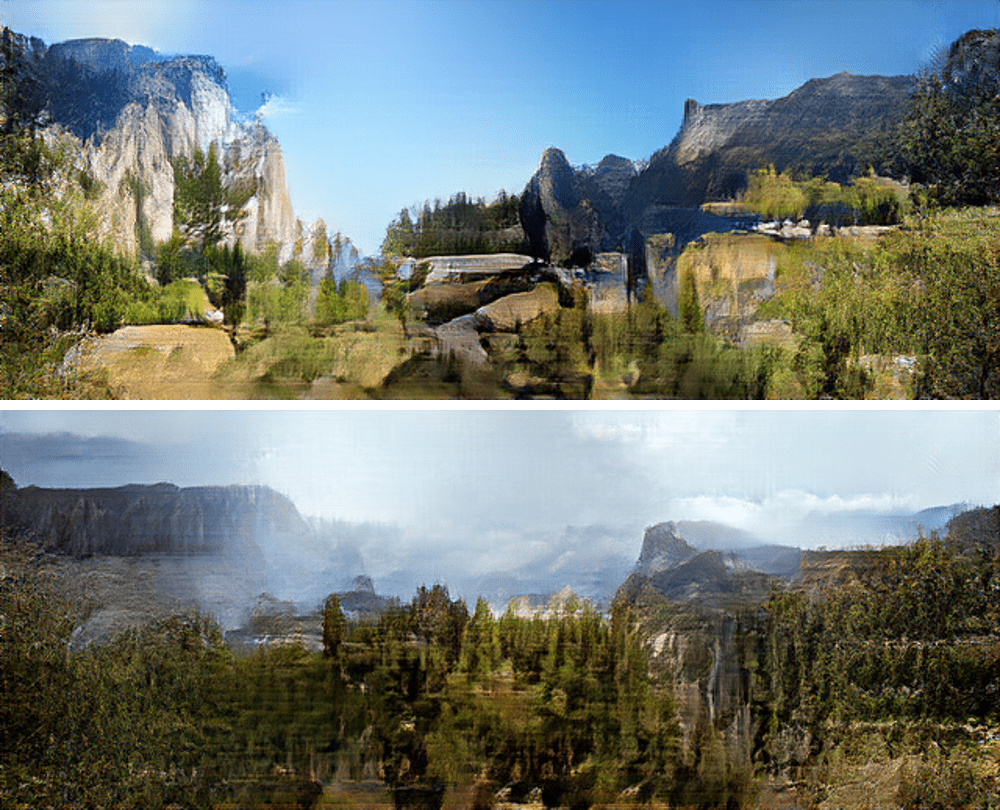} \\
    \midrule
    \includegraphics[height=.40\linewidth]{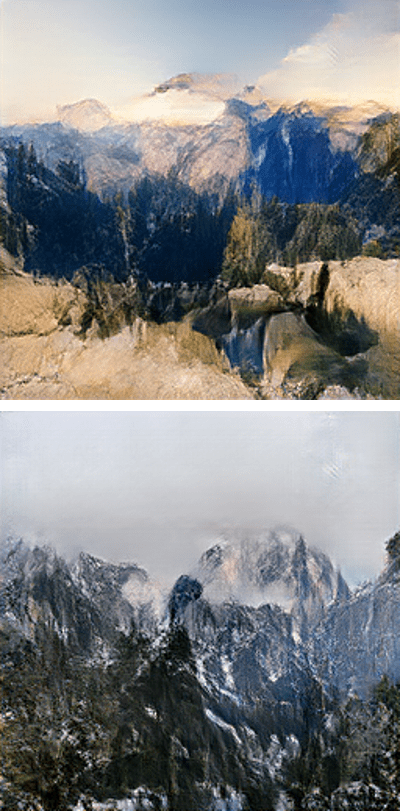} &
    \includegraphics[height=.40\linewidth]{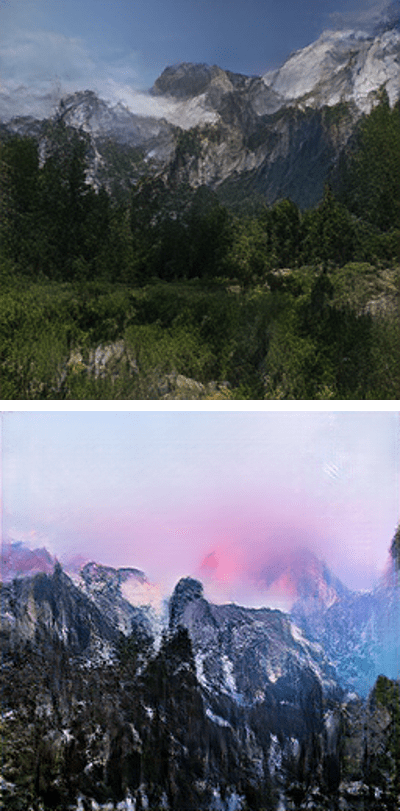} & 
    \includegraphics[height=.40\linewidth]{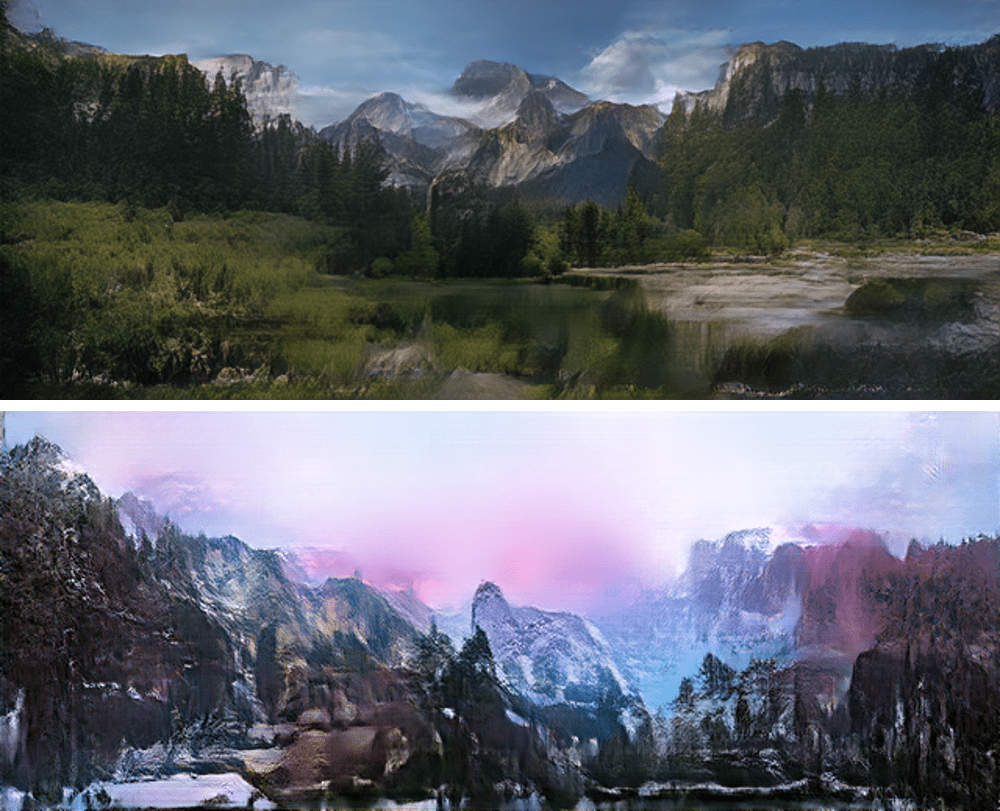} \\
    \end{tabular}
    \caption{\textbf{Unconditional image generation combined with translation and extension.} From top to bottom: summer$\rightarrow$winter, winter$\rightarrow$summer and intra-domain extension.} 
    \label{fig:combination}
\end{figure*}

\begin{figure*}[h]
    \centering
    \setlength\tabcolsep{1.5pt} 
    \begin{tabular}{c:c}
    Input & Diverse extensions \\
    \includegraphics[height=.62\linewidth]{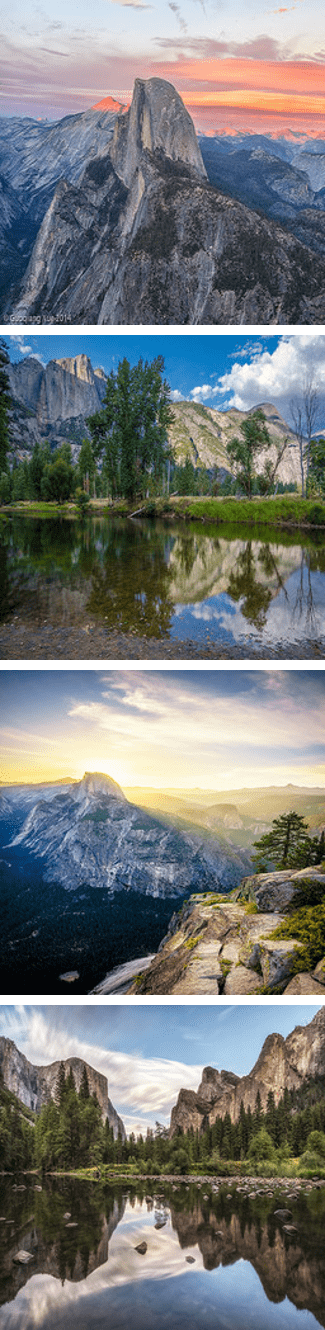} & 
    \includegraphics[height=.62\linewidth]{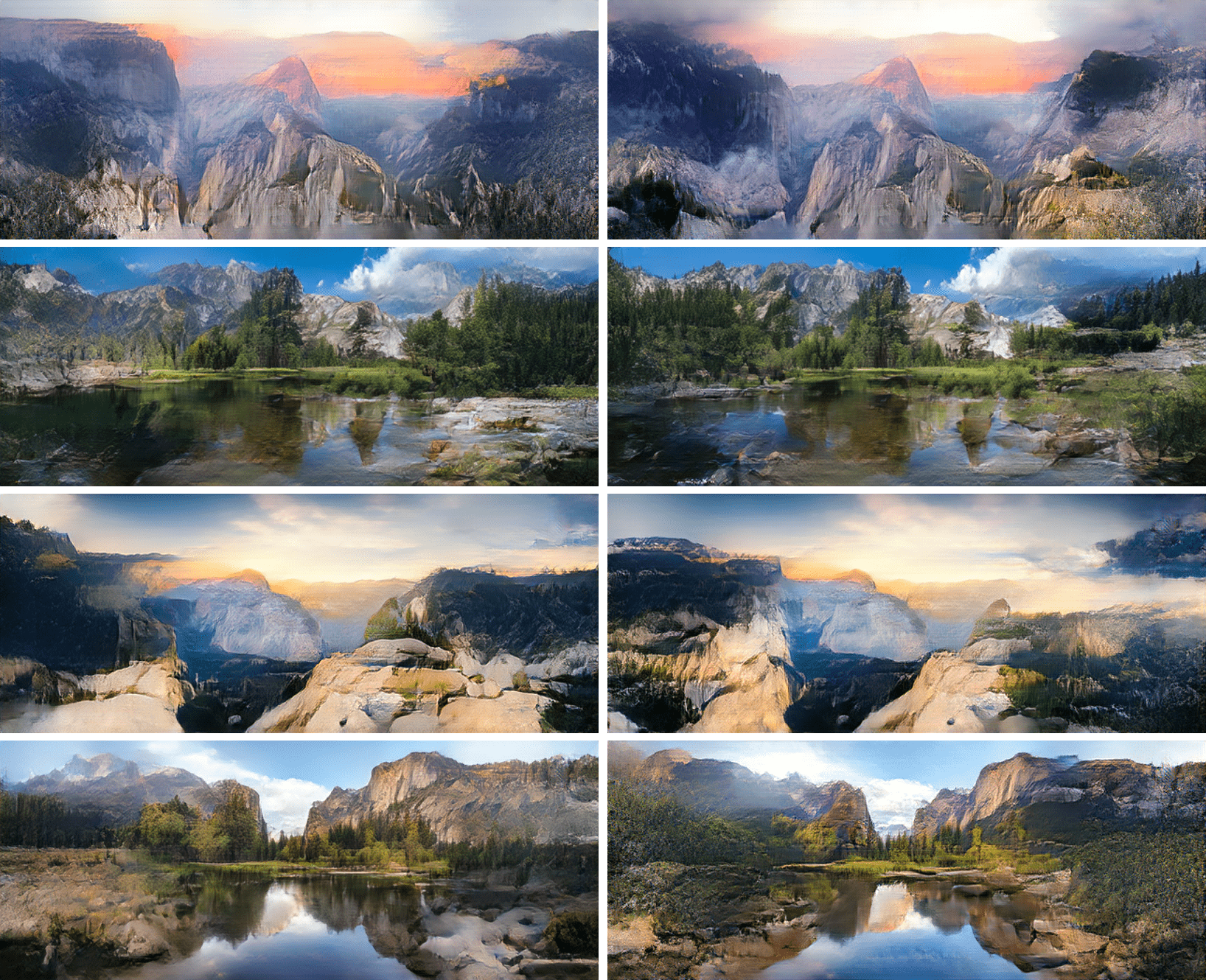} \\
    \includegraphics[height=.62\linewidth]{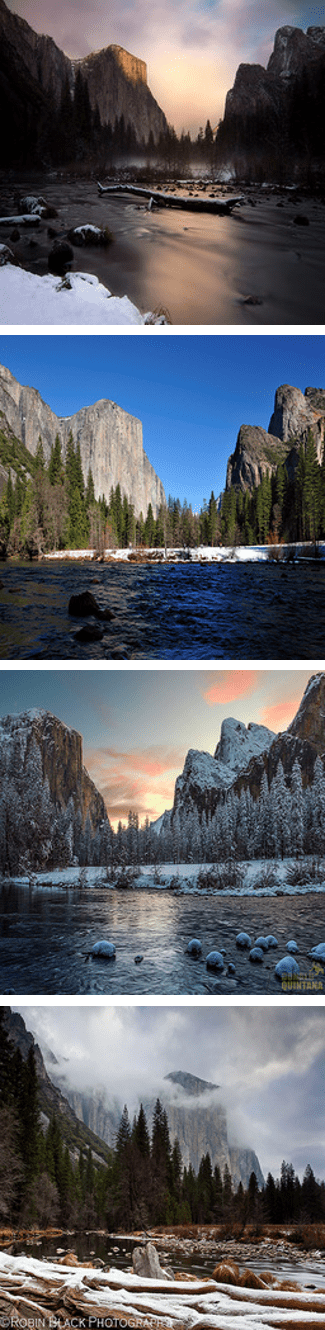} & 
    \includegraphics[height=.62\linewidth]{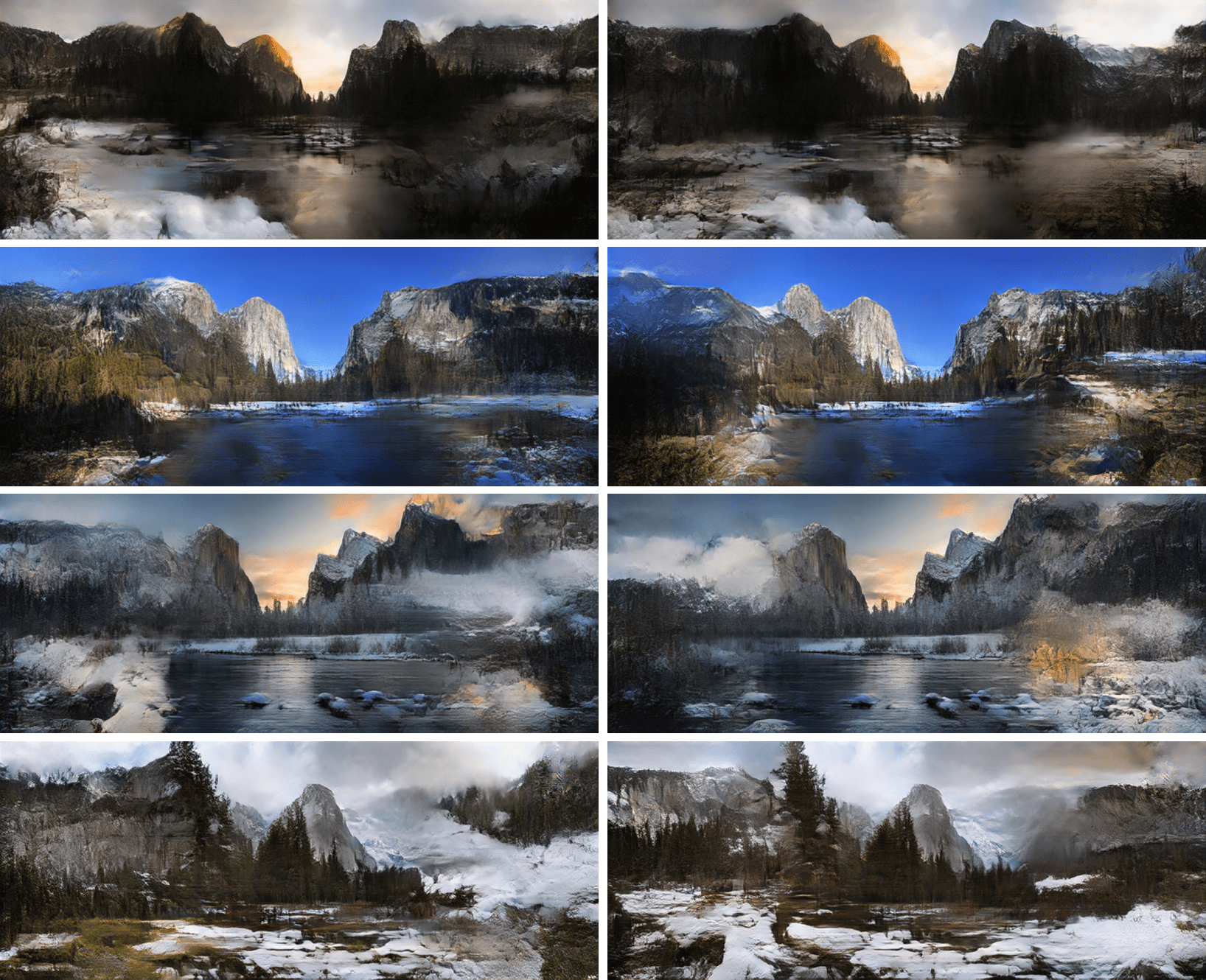} \\
    \end{tabular}
    \caption{\textbf{Diverse image extension.} The transformer is able to generate diverse extension results given an existing image.} 
    \label{fig:img_ext}
\end{figure*}

We demonstrate more application results.
We show the combination of unconditional image generation, image translation, and image extension in Figure~\ref{fig:combination}.
In Figure~\ref{fig:img_ext}, we show the extension on both summer and winter images in Yosemite dataset.
Since the content indices are sampled from the content distribution, the transformer model is able to generate diverse extension results.

\clearpage
%
%
\bibliographystyle{splncs04}
\bibliography{egbib}